\pdfoutput=1
\documentclass{article}

\usepackage[preprint]{neurips_2026}
\usepackage{graphicx}
\usepackage{wrapfig}
\usepackage{subcaption}
\usepackage{bm}

\usepackage[utf8]{inputenc} 
\usepackage[T1]{fontenc}    
\usepackage{hyperref}       
\hypersetup{
    colorlinks=true,
    citecolor=blue,
    linkcolor=blue,
    urlcolor=blue
}
\usepackage{url}            
\usepackage{booktabs}       
\usepackage{amsmath}
\usepackage{amssymb}
\usepackage{tabularx}
\usepackage{longtable}
\usepackage{array}
\usepackage{float}

\usepackage{seqsplit}
\newcolumntype{Y}{>{\raggedright\arraybackslash}X}

\newcommand{\metricb}[1]{\url{#1}}
\usepackage{amsfonts}       
\usepackage{nicefrac}       
\usepackage{microtype}      
\usepackage{xcolor}         

\title{When Plans Change, Remember: Replanning-Guided Visual Memory for Vision-Language-Action Models}

\setcitestyle{aysep={,},citesep={;}}
\usepackage{etoolbox}

\makeatletter

\patchcmd{\@maketitle}
  {\rule{\z@}{24\p@}\@author}
  {\rule{\z@}{12\p@}\@author}
  {}{\PackageWarning{author-spacing}{Author spacing patch failed}}

\makeatother

\author{%
      \normalfont
      \textbf{Victoria Smelova}\textsuperscript{1,*},\quad
      \textbf{Daniil Kazachkov}\textsuperscript{1,*},\quad
      \textbf{Georgiy Kazantsev}\textsuperscript{2}
      \\
      \textbf{Iuliia Tsvetkova}\textsuperscript{3},\quad
      \textbf{Egor Cherepanov}\textsuperscript{1,4,5},\quad
      \textbf{Nikita Kachaev}\textsuperscript{4,5}
      \\
      \textbf{Artem Latyshev}\textsuperscript{1,4,5},\quad
      \textbf{Aleksandr I. Panov}\textsuperscript{1,5},\quad
      \textbf{Alexey K. Kovalev}\textsuperscript{1}
    \\[0.5em]
    \small
    \textsuperscript{1}MIRAI
    \quad
    \small
    \textsuperscript{2}Independent Researcher
    \quad
    \textsuperscript{3}Central University
    \quad
    \small
    \textsuperscript{4}AXXX
    \quad
    \small
    \textsuperscript{5}CogAI Lab
    \\[0.5em]
    \href{https://vika787.github.io/RePlanMem/}{%
    \textbf{RePlanMem Website}%
  }
}

\usepackage[ruled,vlined]{algorithm2e}

\definecolor{RePlanMemBlue}{HTML}{5548C8}

\hypersetup{
    colorlinks=true,
    linkcolor=RePlanMemBlue,
    citecolor=RePlanMemBlue,
    urlcolor=RePlanMemBlue
}

\usepackage{minitoc}

\begin{document}

\doparttoc
\faketableofcontents

\maketitle

\begingroup
\renewcommand{\thefootnote}{\fnsymbol{footnote}}
\footnotetext[1]{These authors contributed equally to this work.}
\endgroup

\vspace{-1.5em}

\begin{abstract}
\vspace{-1em}
Long-horizon, multi-stage Vision-Language-Action (VLA) policies often require
visual information that is no longer present in the current observation.
Retaining the complete observation history is computationally expensive and
highly redundant, motivating selective visual memory. We introduce
\textbf{RePlanMem-VLA} (\emph{Replanning-Guided Keyframe Memory for VLA}), which turns a policy's own plan revisions into an online memory-write signal. Under receding-horizon execution, consecutive action chunks predict some of the same future actions. RePlanMem-VLA treats unusually large disagreement between these temporally aligned predictions as evidence of a task-relevant event. We first compare 35 disagreement measures across single-task autoregressive decoder-only and Flow-Matching Transformers trained from scratch, then validate the selected signals on pretrained OpenVLA-OFT and $\pi_{0.5}$ policies over MIKASA-Robo-VLA tasks. The resulting detectors achieve ROC-AUC of
0.981--0.997. We then use the selected signals in RePlanMem-VLA to trigger updates to a bounded visual memory, with a robot-masked write filter suppressing redundant frames. In the evaluated memory-retention experiments, RePlanMem-VLA preserves task-relevant observations until the later decision point in \(100\%\) of episodes. Real-robot replay data further show that event selection and redundancy suppression extend beyond simulation. These results establish plan revision as a policy-internal signal for controlling what a VLA remembers.
Website: \url{https://vika787.github.io/RePlanMem/}.

\end{abstract}

\par\noindent
\begin{center}
\begin{minipage}{0.88\textwidth}
    \centering
    \vspace{-2em}
    \includegraphics[width=\linewidth]{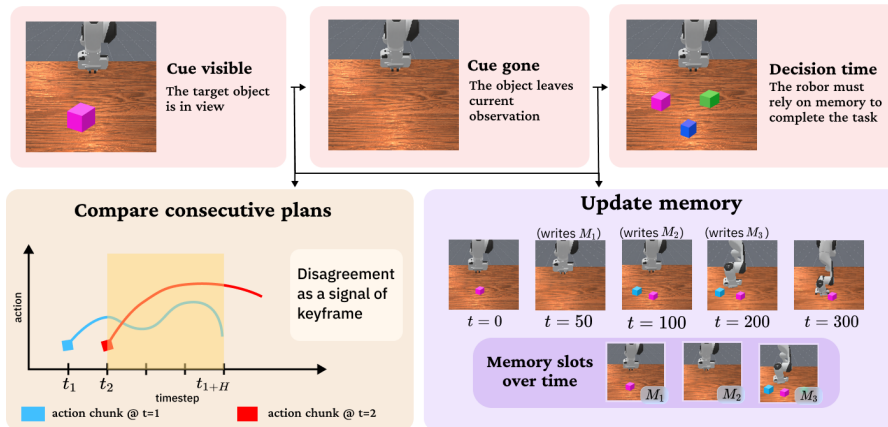}
    \captionsetup{width=\linewidth}
    \captionof{figure}{%
        \textbf{RePlanMem-VLA.}
        Task-relevant cues may disappear from the current observation
        before they are needed. When consecutive replans disagree on
        their overlapping actions, the important observation is stored
        in memory and remains available after the cue is no longer visible.
    }
    \label{fig:visual-abstract}
\end{minipage}
\end{center}

\clearpage

\begin{figure}[t]
    \vspace{-0.5em}
    \centering
    \includegraphics[width=\columnwidth]{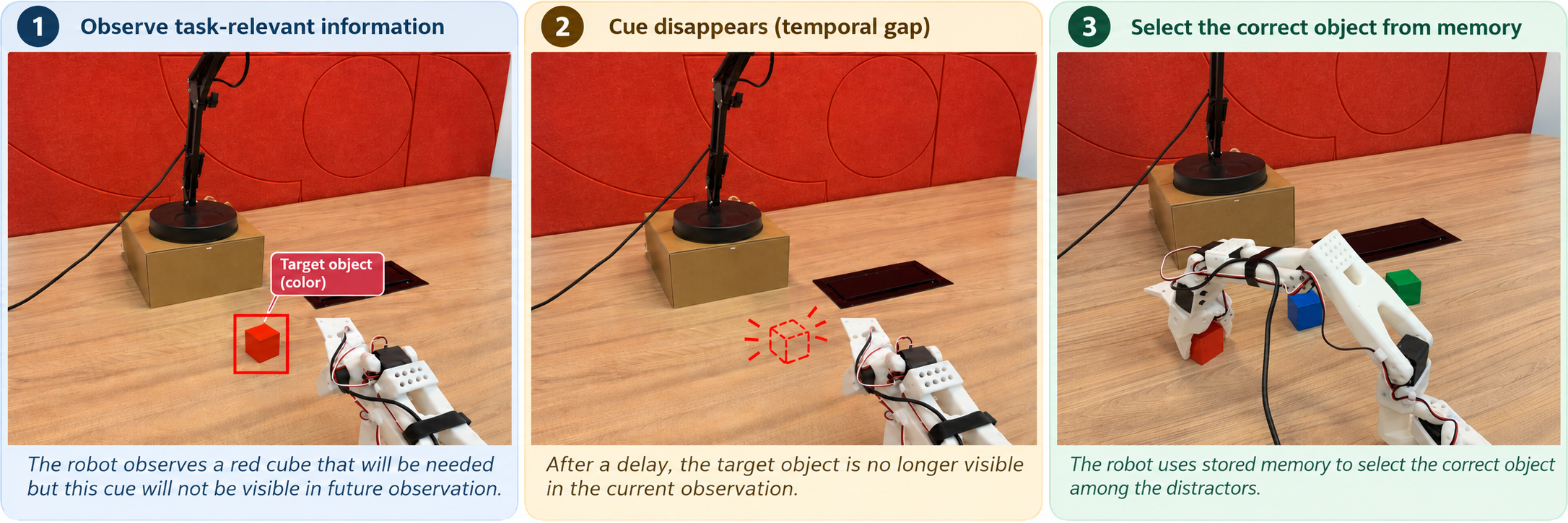}
    \vspace{-1.25em}
    \caption{\textbf{Real-robot delayed-cue setup.}
    The robot observes the target object's color. The cue then leaves
    the observation, and the robot must later select the matching
    object among distractors using stored visual information.
    }
    \label{fig:real-robot-setup}
    \vspace{-1.5em}
\end{figure}

\begin{figure*}[t]
\centering
\includegraphics[width=\textwidth]{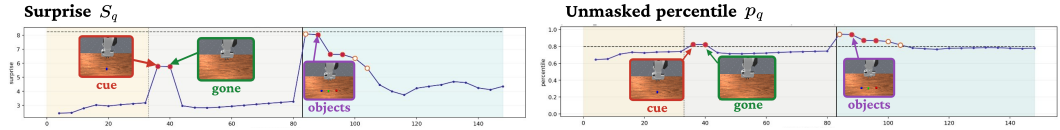}
\caption{
\textbf{Surprise-gated memory updates on the \texttt{RememberColor} task.}
The surprise score \(S_q\) and task-calibrated percentile \(p_q\) identify
candidate update times near transitions between the cue, empty, and action
phases.
}
\label{fig:rq3-writes}
\vspace{-1.5em}
\end{figure*}

\vspace{-2em}
\section{Introduction}
\label{sec:introduction}
\enlargethispage{2\baselineskip}
\vspace{-0.6em}

Long-horizon manipulation often requires  Vision-Language-Action (VLA) policies to use visual information
that has left the current observation, including object properties~\citep{chen2026rmbench,shi2026memoryvla},
earlier environment states~\citep{chen2026rmbench,shi2026memoryvla}, and task
progress~\citep{buamanee2026bihil,feng2026procvlm}.
Retaining the full history preserves these cues but increases context length and
computation while adding visual redundancy~\citep{jang2025contextvla,qiu2026efficientlonghorizonvla,wang2026nativemem}.
We therefore seek a small set of informative observations.

We call these observations \emph{keyframes}: they preserve cues needed
later~\citep{shi2026memoryvla,lei2026robomemarena} or mark task-phase
transitions~\citep{buamanee2026bihil,yi2026criticlooptrisystemvla}. Prior work
selects them with vision-language models~\citep{lei2026robomemarena}, waypoint
heuristics~\citep{shi2023waypoint},
learned retrieval~\citep{sridhar2025memer}, progress estimates~\citep{liu2026palm},
or event detectors~\citep{zeng2026kemo}. We investigate whether the policy's own plan revisions can provide an online signal for selecting these observations. 

Under receding-horizon execution, consecutive action chunks predict some of the same future actions, while the new plan incorporates
the latest observation (Fig.~\ref{fig:visual-abstract}). Comparing these temporally aligned predictions reveals how the policy revises its intended behavior after receiving new visual information.
\textbf{Our central hypothesis is that task-relevant observations induce unusually large revisions in predictions for the same future
actions.} This motivates using plan disagreement as a policy-internal signal to identify candidate keyframes for visual memory.

We introduce \textbf{RePlanMem-VLA}
(\emph{Replanning-Guided Keyframe Memory for VLA}), which derives its selection
signal from revisions between overlapping action predictions already produced
during normal replanning. Selected keyframes become part of the policy's visual context,
keeping past observations available for subsequent action prediction, while a robot-masked write filter suppresses redundant frames, reducing repeated overwrites. At deployment, the trigger requires neither keyframe annotations nor privileged simulator state.

Our experiments are conducted on MIKASA-Robo-VLA~\citep{cherepanov2026mikasarobovlabenchmarkingmemoryvla}, an extension of MIKASA-Robo~\citep{cherepanov2026memory}
for evaluating memory-intensive manipulation with VLA policies. We first screen 35 disagreement metrics using autoregressive decoder-only and
Flow-Matching Transformers~\citep{lipman2023flow}
trained from scratch. We then evaluate detectors based on the selected signals with
pretrained OpenVLA-OFT~\citep{kim2025oft} and
$\pi_{0.5}$~\citep{pi05} policies, obtaining ROC-AUC of
0.981--0.997.
In the evaluated memory-retention experiments, RePlanMem-VLA
preserves task-relevant observations until the later decision
point in \(100\%\) of episodes, while robot-masked filtering
reduces redundant overwrites.
Real-robot replay data further support its event-selection and
redundancy-suppression behavior beyond simulation
(Fig.~\ref{fig:real-robot-setup}).

\vspace{0.5em}
Our main contributions are as follows:

\begin{enumerate}
    \item We introduce \textbf{RePlanMem-VLA}, which identifies keyframes from
    policy-internal revisions of predictions for the same future actions across
    consecutive replanning steps (Section~\ref{sec:plan-revision-signal}).
    
    \item We \textbf{identify disagreement measures that best align with keyframe events} through systematic analysis across decoder-only and Flow-Matching Transformer policies, and verify their transfer to pretrained OpenVLA-OFT and $\pi_{0.5}$ policies
    (Sections~\ref{sec:rq1-metric-screening} and~\ref{sec:rq2-vla-keyframe-detection}).
     
     \item We develop a \textbf{bounded visual-memory mechanism} where plan disagreement determines when to write and redundancy-aware filtering, including robot-masked clustering, determines what to retain, substantially reducing redundant updates while preserving task-relevant observations (Sections~\ref{sec:method} and~\ref{sec:rq3}).
    
\end{enumerate}

\begin{figure*}[t]
    \centering
    \includegraphics[width=\textwidth]{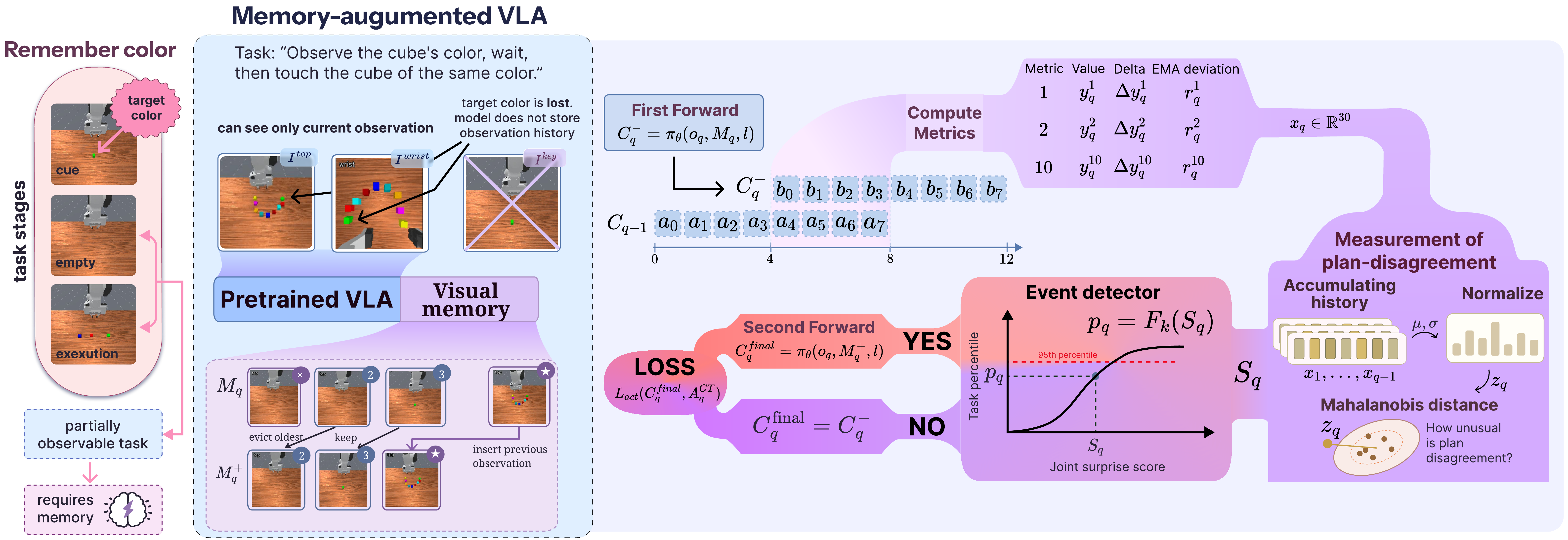}
    \vspace{-0.5em}
    \caption{\textbf{RePlanMem-VLA overview.}
    At replanning step \(q\), the VLA first predicts a chunk
    \(C_q^{-}\) from the current observation, instruction, and existing
    memory. The temporally aligned overlap with the preceding plan
    is converted into a task-calibrated
    surprise percentile \(p_q\).
    If \(p_q>\tau\), a selected observation is
    written to FIFO memory and the policy performs a second forward pass with the updated memory, yielding \(C_q^{\mathrm{final}}=C_q^{+}\),
    otherwise \(C_q^{\mathrm{final}}=C_q^{-}\).}
    \label{fig:keyframe-memory-method}

    \vspace{-0.5em}
\end{figure*}

\vspace{-2em}
\section{Related Work}
\label{sec:related}
\enlargethispage{1\baselineskip}
\vspace{-0.5em}

\begingroup
\makeatletter
\renewcommand{\paragraph}{%
  \@startsection{paragraph}{4}{\z@}%
  {0.5ex plus 0.2ex minus 0.1ex}%
  {-1em}%
  {\normalfont\normalsize\bfseries}%
}
\makeatother

\paragraph{Memory for long-horizon, partially observable policies.}
Long-horizon manipulation often requires access to information from earlier in
the episode. Recent VLA work addresses this through
recurrent or belief-state
latents~\citep{cherepanov2026muvla,li2026rememvla},
explicit memory representations
~\citep{shi2026memoryvla,guo2025chameleon}, long-context or
compressed multi-frame conditioning
~\citep{jang2025contextvla}, and retrieval- or structure-based memory
~\citep{sridhar2025memer,haresh2026notestoself,
chung2025objectcentric,li2026spatialmem}.
Memory-dependent manipulation is also studied in dedicated benchmarks
~\citep{cherepanov2026mikasarobovlabenchmarkingmemoryvla,chen2026rmbench,lei2026robomemarena}. These works study how historical information is represented or retrieved, whereas our focus is on when an observation should be admitted to a bounded memory.

\paragraph{Keyframe and event selection for visual memory.}
Existing approaches identify important observations using motion or
bottleneck-pose heuristics~\citep{james2022qattention},
learned frame scoring~\citep{mark2025bpp,chen2025keyframechaining},
progress, value, or critic signals
~\citep{liu2026palm,yi2026criticlooptrisystemvla,buamanee2026bihil},
trajectory annotation~\citep{feng2026procvlm}, or dedicated event detectors
~\citep{zeng2026kemo} --- each
relying on a separate selector or external signal. We instead derive the selection signal directly from changes in the acting policy's own predicted action plans.

\paragraph{Action chunking, replanning, and plan consistency.}
Action-chunking policies predict short action sequences and under receding-horizon
execution successive chunks overlap on predictions for the same future actions
~\citep{zhao2023act,chi2023diffusionpolicy,kim2025oft,pi05}. Prior work mainly uses this overlap to improve consistency
~\citep{zhao2023act,liu2024bid}
or studies the trade-off between open-loop execution and replanning
~\citep{lazzati2026doesactionchunkingimprove,zeng2026revisitingopenloop,
chen2026dreamchunk,jing2025mixturehorizons}.
We instead use disagreement between overlapping predictions as a signal that the
new observation has changed the policy's intended behavior.

\par
\endgroup

Memory representation, keyframe selection, and plan consistency are typically
studied separately, whereas we connect them by using plan disagreement as a
policy-internal signal for memory updates.
 
\vspace{-2em}

\section{RePlanMem-VLA}
\label{sec:method}

\subsection{Overview}
%
In long-horizon manipulation, task-relevant visual information often leaves the cameras before it is needed.  A policy that sees only the current frame cannot recover that cue, so we concatenate the live views with a bounded visual memory of \(B\) slots. 
Slots are empty at the start of an episode and are never carried across episode boundaries. 
The policy predicts a chunk of \(K\) actions from one observation and executes only the first \(H<K\) before replanning, so consecutive chunks overlap on  \(K{-}H\) future steps: the same intended actions, predicted before and after the new frame. 
Prior work treats disagreement on that overlap as noise to suppress.
We read a large revision as evidence that the new observation was task-relevant and use it to trigger a memory-update proposal. The direct update rule stores the immediately preceding frame, whereas the redundancy-aware variant retains only visual states that are not already represented in memory.
Figures~\ref{fig:visual-abstract} and~\ref{fig:keyframe-memory-method} and Algorithm~\ref{alg:surprise-memory} summarize the procedure. 
At every query the policy first predicts a pre-write chunk from the memory available before a potential write, then overlap disagreement is converted, without gradients, into a task-calibrated surprise score and an event triggers frame selection and a memory update only when a visually novel state is found. The final action prediction is conditioned on the resulting memory state, and only this prediction receives the original supervised fine-tuning (SFT) objective.  Event detection never sees expert actions, so memory writes are driven only by the policy's own plan revisions.

\subsection{Memory-augmented action policy}
In partially observable manipulation, identical current observations
can require different actions depending on earlier cues or task
progress~\citep{cherepanov2026memory,chen2026rmbench}, as illustrated by the delayed-cue setup in Fig.~\ref{fig:real-robot-setup}.
A policy conditioned only on the current observation cannot distinguish such histories.
Historical context can resolve this ambiguity, motivating multi-frame
conditioning and memory-augmented
policies~\citep{jang2025contextvla,shi2026memoryvla}.  To bound the amount of historical visual input, we retain at most \(B\) observations. The resulting online selection problem is which observations
to preserve for future control. We call these observations keyframes, focusing on transient cues needed later and observations marking changes in task phase.

 Let \(q\) index policy queries, \(o_q\) denote the current multi-view observation, and \(\ell\) denote the language instruction. Action chunking predicts multiple future controls from a single
observation, supporting temporally coordinated behavior and efficient
execution~\citep{zhao2023act,chi2023diffusionpolicy,kim2025oft}.
The policy predicts a chunk of \(K\) actions \(C_q=(a_{q,0},\ldots,a_{q,K-1})\) but executes only the first \(H<K\) of them before replanning. We refer to \(K\) and \(H\) as the \emph{chunk horizon} and \emph{execution stride}. Here, \(a_{q,j}\in\mathbb{R}^{d}\) denotes the action predicted
at query \(q\) for the \(j\)-th step of the chunk.

The visual memory before query \(q\) is an ordered collection of at most \(B\) keyframes \(M_q=(M_q^1,\ldots,M_q^{m_q})\), \(m_q\leq B\), whose slots are appended to the policy's visual context alongside the current camera views: with a top and a wrist camera the visual input is \(X_q=[\,I_q^{\mathrm{top}},I_q^{\mathrm{wrist}},M_q^1,\ldots,M_q^{m_q}\,]\).

All slots are empty at the beginning of an episode, and memory state is never carried across episode boundaries. 
At query \(q\), a first forward pass uses the memory state that existed before the current event decision and yields the \emph{pre-write plan} 
\(C_q^-=\pi_\theta(o_q,M_q,\ell)\), which provides both a candidate control output and the policy-internal signal used to decide whether memory should be updated. We denote the chunk used for execution after the memory-update decision by \(C_q^{\mathrm{final}}\).

\subsection{Plan revision as an event signal}
\label{sec:plan-revision-signal}
Two consecutive action chunks make predictions for some of the same future time steps. 
Because \(H\) actions of the preceding chunk have already been executed, the aligned overlap has length \(K-H\). 
We therefore compare
\begin{equation}
    U_q = C_{q-1}^{\mathrm{final}}[H:K],
    \qquad
    V_q = C_q^-[0:K-H].
    \label{eq:chunk-overlap}
\end{equation}
Both sequences estimate the same future actions, but \(V_q\) is conditioned on the newly observed scene, so a large discrepancy indicates that new visual evidence has caused the policy to revise its intended behavior. 
We summarize this revision with \(D\) complementary detector signals
\(\bm y_q=[y_q^1,\ldots,y_q^D]^\top\). These signals include aligned
action error, directional disagreement, temporally localized deviations,
translation, rotation, and gripper changes. The deployed detector also
includes the distance between consecutive detached top-view VLA features as one of these signals.
In our configuration \(D=10\).

To distinguish persistent model uncertainty from abrupt changes, we augment
these values with their temporal differences
\(\Delta\bm y_q=\bm y_q-\bm y_{q-1}\) and with causal deviations from an
exponential moving average (EMA),
\[
    \widetilde{\bm m}_q
    =
    \beta\widetilde{\bm m}_{q-1}
    +(1-\beta)\bm y_q,
    \qquad
    \bm d_{q,\mathrm{EMA}}
    =
    \bm y_q-\widetilde{\bm m}_{q-1}.
\]
The complete detector input is
\begin{equation}
    \bm x_q =
    \left[
        \bm y_q^\top,
        \Delta\bm y_q^\top,
        \bm d_{q,\mathrm{EMA}}^\top
    \right]^\top
    \in\mathbb R^{3D},
    \label{eq:event-features}
\end{equation}
which is \(\mathbb R^{30}\) in our configuration. All quantities entering
Eqs.~\eqref{eq:chunk-overlap} and~\eqref{eq:event-features} are computed
from stop-gradient copies of action chunks and visual features.
Consequently, the event detector cannot shape the policy by differentiating
through its decision rule.

\subsection{Causal, task-conditioned event detection}
The components of \(\bm x_q\) have different units and dynamic ranges, so we
standardize each of them using statistics available strictly before the current
query, \(z_q^i=(x_q^i-\mu_{q-1}^{i,(k)})/(\sigma_{q-1}^{i,(k)}+\varepsilon)\).
The current sample is scored before it is incorporated into the running
statistics.  All detector statistics --- the moments, the matrix below, and the
reference cumulative distribution function (CDF) of Eq.~\eqref{eq:event-rule} --- are maintained separately for each
task \(k\), so that no cross-task scale difference enters the score.  Let
\(\bm\Sigma_{q-1}^{(k)}\) be the causal second-moment matrix of the normalized
features. Because the preceding normalization is component-wise, its diagonal
is unity up to the lag of the running moments, so \(\bm\Sigma_{q-1}^{(k)}\)
acts as a correlation matrix.  The joint deviation of the features is measured
by the regularized Mahalanobis surprise:
\begin{equation}
    S_q =
    \sqrt{
        \bm z_q^\top
        \left(\bm\Sigma_{q-1}^{(k)}+\lambda\bm I\right)^{-1}
        \bm z_q
    },
    \label{eq:mahalanobis}
\end{equation}
where \(\lambda>0\) stabilizes the inversion. We set it to a fixed fraction of the mean eigenvalue of \(\bm\Sigma_{q-1}^{(k)}\), which makes the regularizer scale-free across tasks.  Unlike an unweighted Euclidean norm, Eq.~\eqref{eq:mahalanobis} accounts for correlations among different revision signals.

Natural plan variability differs across tasks in multi-task SFT, which is why the calibration above is per-task while the decision threshold is shared. Writing \(F_k(s)=\Pr(S\leq s\mid k)\) for the empirical CDF of past surprise scores for task \(k\), the event percentile and binary decision are
\begin{equation}
    p_q = F_k(S_q),
    \qquad
    e_q = \mathbf{1}\!\left[p_q>\tau\right].
    \label{eq:event-rule}
\end{equation}

\begin{wrapfigure}{r}{0.5\textwidth}
\vspace{-2.0em}
\centering
\includegraphics[width=\linewidth]{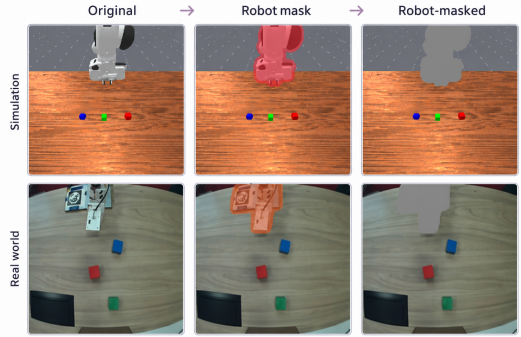}
\vspace{-2em}
\caption{%
\textbf{Examples of robot masking in simulation and real-robot observations.}
Each example shows the original observation, the estimated robot mask,
and the masked image used for redundancy comparison.
The policy receives the original RGB frame rather than the masked image.
}
\label{fig:rq3-masks}
\vspace{-3em}
\end{wrapfigure}
Thus \(\tau=0.99\), for example, assigns the same interpretation --- an event more surprising than \(99\%\) of the task-specific reference distribution --- across heterogeneous tasks.  At the beginning of SFT, memory writes are disabled for a short warm-up period (the first \(5\)--\(10\%\) of training), while the policy is optimized normally and detector statistics are collected.  After warm-up, the CDFs and causal moments can be updated with rolling histories to track the changing policy. Once training is complete, these statistics are frozen and kept fixed at inference time to preserve a stable detector calibration. The first query in each episode has no preceding chunk and cannot produce a plan-overlap event.

\subsection{Redundancy-aware frame selection}
\label{sec:redundancy-aware-selection}

An event determines \emph{when} memory should be reconsidered, but not
necessarily which frame should be stored. At an event, we jointly consider
the current FIFO contents and a short causal window \(W_q\) of observations
preceding the current query. We use the three observations preceding the
event query (\(W=3\)) and apply deterministic complete-link clustering with
threshold \(\delta=0.01\). From each novel cluster, the most recent candidate
is selected as its representative. The current observation is excluded so that
selection cannot use information obtained after the detected plan revision.

We group visually similar frames using complete-link clustering and treat a
candidate cluster as novel only if it contains no frame already represented
in memory. From each novel cluster, we retain its most recent candidate and
append the selected representatives to the bounded FIFO in chronological
order. Thus, one detector event may result in zero, one, or multiple writes.
Visual masks, when available, are used only to suppress irrelevant appearance
changes during comparison, while the original RGB frames are stored and supplied
to the policy. The direct update rule is the special case in which the
selector returns only the immediately preceding frame without redundancy
filtering.

\par\noindent
\begin{minipage}{\linewidth}
\begin{algorithm}[H]
\SetAlgoLined
\DontPrintSemicolon
\SetKwComment{Comment}{// }{}
\SetKwInOut{Input}{Input}
\SetKwInOut{Output}{Output}
\caption{Surprise keyframe memory at query \(q\).}
\label{alg:surprise-memory}

\Input{Observation \(o_q\), instruction \(\ell\), task id \(k\), memory
\(M_q\) with \(B\) slots, previous chunk \(C_{q-1}^{\mathrm{final}}\), chunk
horizon \(K\), execution stride \(H\), percentile threshold \(\tau\), causal
statistics \((\bm\mu,\bm\sigma,\bm\Sigma,F_k)_{q-1}\), candidate-window size
\(W=3\), clustering threshold \(\delta=0.01\).}

\Output{Final action chunk \(C_q^{\mathrm{final}}\),
updated memory \(M_{q+1}\).}

\Comment{1. pre-write forward pass}
\(C_q^-\leftarrow\pi_\theta(o_q,M_q,\ell)\)\;

\If{\(q\) is the episode's first query}{
  \(M_q^+\leftarrow M_q\)\;
  \(C_q^{\mathrm{final}}\leftarrow C_q^-\)\;
}
\Else{
  \Comment{2. plan-revision features (stop-gradient)}
  \(U_q\leftarrow C_{q-1}^{\mathrm{final}}[H{:}K]\),\;
  \(V_q\leftarrow C_q^-[0{:}K{-}H]\)\;
  \(\bm x_q\leftarrow
  [\bm y_q,\Delta\bm y_q,\bm d_{q,\mathrm{EMA}}]\)
  by Eq.~\eqref{eq:event-features}\;

  \Comment{3. causal scoring, then statistics update}
  \(\bm z_q\leftarrow
  (\bm x_q-\bm\mu_{q-1})
  \oslash(\bm\sigma_{q-1}+\varepsilon)\)\;
  \(S_q\leftarrow\) Mahalanobis\((\bm z_q)\)
  by Eq.~\eqref{eq:mahalanobis}\;
  \(p_q\leftarrow F_k(S_q)\);
  \(e_q\leftarrow\mathbf 1[p_q>\tau]\)\;
  update \((\bm\mu,\bm\sigma,\bm\Sigma,F_k)\)
  with \(\bm x_q\)\;

  \Comment{4. event-gated frame selection and memory update}
  \(R_q\leftarrow\varnothing\)\;

  \If{\(e_q=1\) and warm-up has ended}{
    \(Q_q\leftarrow
    \{I_{q-W}^{\mathrm{top}},\ldots,I_{q-1}^{\mathrm{top}}\}\)\;
    \(Z_q\leftarrow M_q\cup Q_q\)\;
    cluster \(Z_q\) using deterministic complete-link clustering
    with threshold \(\delta\)\;

    \ForEach{cluster containing no frame from \(M_q\)}{
      append the most recent candidate to \(R_q\)\;
    }

    sort \(R_q\) chronologically\;
  }

  \eIf{\(R_q\neq\varnothing\)}{
    \(M_q^+\leftarrow
    \operatorname{FIFOAppend}(M_q,R_q)\)\;
    \(C_q^{\mathrm{final}}\leftarrow
    \pi_\theta(o_q,M_q^+,\ell)\)\;
  }{
    \(M_q^+\leftarrow M_q\);
    \(C_q^{\mathrm{final}}\leftarrow C_q^-\)\;
  }
}

\Comment{5. execution and (training only) supervision}
Execute \(C_q^{\mathrm{final}}[0{:}H]\);
\(M_{q+1}\leftarrow M_q^+\)\;
Accumulate
\(\mathcal L_{\mathrm{act}}
(C_q^{\mathrm{final}},A_q^{\mathrm{GT}})\);
no gradient flows through \(\bm x_q\),
the statistics, or \(e_q\)\;

\end{algorithm}
\end{minipage}
\par\vspace{-0.5em}
\noindent

\subsection{Event-gated memory update and second forward pass}
\label{sec:event-gated-memory}
\vspace{-0.5em}
When an event is detected at query \(q\), the frame-selection rule returns a
possibly empty sequence \(R_q\) of candidate keyframes. The direct rule returns
the immediately preceding top-view observation, whereas the redundancy-aware
rule may return zero, one, or multiple frames. A fixed-capacity FIFO operator
produces
\[
    M_q^+ = \operatorname{FIFOAppend}(M_q,R_q),
\]
inserting the selected frames in chronological order and discarding the
oldest items when all \(B\) slots are occupied. If \(R_q=\varnothing\),
memory remains unchanged even though the detector emitted an event.

At inference time, an actual memory update is followed by a second evaluation
of the same policy on the current observation and instruction,
\[
    C_q^{\mathrm{final}}
    =
    \pi_\theta(o_q,M_q^+,\ell).
\]
If no frame is written, no additional inference pass is required and
\(C_q^{\mathrm{final}}=C_q^-\). During distributed SFT, the inference-only
pre-write pass is followed by one differentiable final pass for the complete
micro-batch, so rows without a memory update simply use the unchanged memory
state. This keeps the distributed computation graph synchronized when event
counts differ across ranks.

\vspace{-0.5em}
\subsection{Supervised fine-tuning objective}
\vspace{-0.5em}
Let \(A_q^{\mathrm{GT}}\) be the expert action chunk associated with query \(q\).  We retain the backbone's standard action objective and apply it only to
the selected final chunk, \(\mathcal L_{\mathrm{SFT}}(\theta)=\sum_q\mathcal L_{\mathrm{act}} (C_q^{\mathrm{final}},A_q^{\mathrm{GT}})\).  Gradients flow through the policy invocation that produced \(C_q^{\mathrm{final}}\), but not through the detector, its running statistics, or the discrete memory-write decision. 
Expert actions supervise control only: they are not used to identify events or select keyframes. 
The same causal detector and memory-update procedure can therefore be used at inference time without keyframe annotations or privileged simulator state.

 \vspace{-0.5em}

\section{Experimental evaluation}
\label{sec:exps}
\vspace{-0.5em}

Our experiments address three Research Questions (RQs):

\begin{enumerate}
    \item[\textbf{RQ1.}] Does revision of overlapping action chunks contain
    a task-relevant signal, and which disagreement measures extract it most
    reliably?

    \item[\textbf{RQ2.}] Does this signal transfer to pretrained VLA policies
    across different backbones and replanning frequencies, and can it support
    accurate online event detection?

    \item[\textbf{RQ3.}] Can the resulting surprise signal be used to drive
    selective visual-memory updates during VLA fine-tuning?
\end{enumerate}

\vspace{-0.5em}
\subsection{RQ1: Disagreement signals in single-task policies}
\label{sec:rq1-metric-screening}
\vspace{-0.5em}

We first examine whether disagreement between overlapping action predictions contains task-relevant information using single-task autoregressive and Flow-Matching Transformers trained from scratch. This analysis allows us to efficiently screen a
large set of disagreement metrics at substantially lower computational cost before transferring the most informative signals to pretrained VLA
policies. We evaluate 35 disagreement metrics
(Appendix~\ref{app:all-metrics}) across eight MIKASA-Robo-VLA tasks (Appendix~\ref{app:rq1-setup}) and examine
whether their behavior remains consistent across different task structures and amounts of available historical information. For each architecture, we compare
two regimes: full-context, where the policy has
access to the full observation history up to the current query, and current-only, where the prediction is conditioned only on the current observation.

Visual analysis of disagreement profiles shows that scores often
remain near a relatively stable baseline, with peaks occurring near
keyframes. This pattern motivates a simple threshold-based rule:
a frame is predicted as a keyframe when its disagreement score
exceeds a threshold.
We use Frame ROC-AUC as the primary screening criterion,
evaluating how well these predictions match reference keyframe labels across detection thresholds. A more detailed analysis, including correlation analysis, and task-specific results, is provided in
Appendix~\ref{app:rq1-analysis}.

Experiments show that disagreement measures remain informative both with and without observation history, although their effectiveness varies across tasks and settings.
On tasks with clear phase transitions, such as
\texttt{RememberColor3-VLA-v0} and \texttt{ShellGameColorLampTouch-VLA-v0}, for autoregressive policies, Causal Z-score reaches Frame ROC-AUC
of \(0.798 / 0.746\) under full-context/current-only conditioning, with EMA deviation showing a similar pattern (\(0.790 / 0.712\)).
For Flow-Matching policies, distributional measures achieve higher scores: Wasserstein-1 reaches  \(0.9388\) and Energy distance
\(0.9427\). Translation, rotation, and gripper disagreement also perform well,
with scores ranging from \(0.851\) to \(0.911\). Causal Z-score remains stable
(\(0.696/0.690\)), showing that the causal temporal signal persists across both action-generation paradigms.

The broader analysis establishes disagreement between consecutive action chunks as a useful signal of keyframes, with several types of plan revision measures contributing useful information
across tasks. These signals remain informative across action-generation architectures and
context regimes and remain visible on failed trajectories (Appendix~\ref{app:rq1-outcomes}). Correlation analysis further reveals substantial redundancy within
metric families (Appendix~\ref{app:rq1-correlations}), indicating
that closely related measures often provide similar information about plan-revision structure. These findings motivate selecting a compact set of complementary
disagreement features for the subsequent VLA experiments.

\vspace{-1.5em}
\subsection{RQ2: Event detection with pretrained VLA policies}
\label{sec:rq2-vla-keyframe-detection}

\begin{wraptable}{l}{0.42\textwidth}
\centering
\footnotesize
\setlength{\tabcolsep}{0pt}
\renewcommand{\arraystretch}{1.05}
\vspace{-1.5em}
\caption{\textbf{Detector features used in RQ2.}}
\label{tab:rq1-selected-features}
\vspace{-1em}
\begin{tabularx}{\linewidth}{@{}Y@{}}
\toprule

\textbf{Overlap L2}\newline
Mean L2 distance between aligned action predictions. \\
\addlinespace[3pt]

\textbf{BiD lower bound}\newline
Weighted sum of L2 distances. \\
\addlinespace[3pt]

\textbf{Causal Z-score}\newline
First-action revision relative to its causal baseline. \\
\addlinespace[3pt]

\textbf{CUSUM}\newline
Accumulated one-sided temporal change. \\
\addlinespace[3pt]

\textbf{EMA deviation}\newline
Deviation of the first-action change from its causal
exponential moving average. \\
\addlinespace[3pt]

\textbf{Translation cosine distance}\newline
Mean cosine distance.\\
\addlinespace[3pt]

\textbf{Translation disagreement}\newline
Difference between translation predictions
in controller-physical space.  \\
\addlinespace[3pt]

\textbf{Rotation disagreement}\newline
Geodesic distance between predicted rotations in \(SO(3)\). \\
\addlinespace[3pt]

\textbf{Gripper disagreement}\newline
Normalized difference between gripper predictions. \\

\midrule

\textbf{VLM top-view L2}\newline
L2 distance between top-view visual features
from consecutive observations. \\

\bottomrule
\end{tabularx}
\vspace{-3.5em}
\end{wraptable}
We next evaluate whether a compact combination of metrics remains effective
for online event detection in pretrained VLA policies across different
backbones and replanning frequencies. Building on RQ1, we select nine representative metrics spanning the distributional, cumulative, and causal temporal signals and augment them with a VLA-specific visual feature that captures
changes in the policy's visual representation between consecutive
observations. The resulting ten features
(Table~\ref{tab:rq1-selected-features}) serve as input to learned
keyframe detectors.

We evaluate pretrained OpenVLA-OFT~\citep{kim2025oft} and $\pi_{0.5}$~\citep{pi05} policies on 19 MIKASA-Robo-VLA tasks, under two execution regimes: replanning after every action
(RH1, \(H=1\)) and executing four actions before replanning
(EXEC4, \(H=4\)). Both policies predict chunks of \(K=8\) actions.
For each pair of consecutive plans, we compute the selected
metrics over their temporally aligned overlap.
Together with the visual-change feature, these measurements
provide the input for three types of keyframe detectors:
XGBoost~\citep{chen2016xgboost},
CatBoost~\citep{prokhorenkova2018catboost}, and a causal
LSTM~\citep{hochreiter1997lstm}.
The detectors are trained using simulator phase transitions
as reference keyframe labels.

\begin{wraptable}{r}{0.55\textwidth}
  \vspace{-1.5em}
  \centering
  \caption{\textbf{Keyframe detection under RH1 and EXEC4.}}
  \label{tab:vla-rh-exec4}
  \vspace{-0.4em}
  \scriptsize
  \setlength{\tabcolsep}{2.5pt}
  \renewcommand{\arraystretch}{1.05}
  \resizebox{\linewidth}{!}{%
  \begin{tabular}{llccc}
    \toprule
    \textbf{Policy} & \textbf{Regime} &
    \textbf{Classifier} & \textbf{PR-AUC} & \textbf{ROC-AUC}  \\
    \midrule
    OpenVLA-OFT & RH1  & XGBoost  & 0.932 & 0.982 \\
    OpenVLA-OFT & RH1  & CatBoost & 0.929 & 0.981  \\
    OpenVLA-OFT & RH1  & LSTM     & 0.962 & 0.989 \\
    OpenVLA-OFT & EXEC4 & XGBoost  & 0.996 & 0.997 \\
    OpenVLA-OFT & EXEC4 & CatBoost & 0.995 & 0.996 \\
    OpenVLA-OFT & EXEC4 & LSTM     & 0.984 & 0.990  \\
    \midrule
    $\pi_{0.5}$ & RH1   & XGBoost  & 0.928 & 0.984 \\
    $\pi_{0.5}$ & RH1  & CatBoost & 0.924 & 0.984  \\
    $\pi_{0.5}$ & RH1   & LSTM     & 0.933 & 0.985  \\
    $\pi_{0.5}$ & EXEC4   & XGBoost  & 0.982 & 0.988  \\
    $\pi_{0.5}$ & EXEC4  & CatBoost & 0.981 & 0.987 \\
    $\pi_{0.5}$ & EXEC4   & LSTM     & 0.983 & 0.988 \\
    \bottomrule
  \end{tabular}%
  }
  \vspace{-1em}
\end{wraptable}

\vspace{1em}
For detection evaluation we use ROC-AUC and PR-AUC.
As in RQ1, ROC-AUC measures how well detector scores separate
keyframes from other observations across thresholds.
PR-AUC evaluates the trade-off between precision and recall,
reflecting whether the detectors identify keyframes while
limiting false detections. Table~\ref{tab:vla-rh-exec4} shows that both metrics remain high
across policies, execution regimes, and detector types:
ROC-AUC ranges from \(0.981\) to \(0.997\), and PR-AUC from
\(0.924\) to \(0.996\).
The causal LSTM, which also incorporates information from
preceding queries, achieves the highest PR-AUC under RH1
for both OpenVLA-OFT (\(0.962\)) and $\pi_{0.5}$ (\(0.933\)).
Detection remains strong under EXEC4, where XGBoost achieves
the highest PR-AUC for OpenVLA-OFT (\(0.996\)), while the
LSTM performs best for $\pi_{0.5}$ (\(0.983\)).

We additionally evaluate \(\mathrm{event\_recall@1}\),
which measures whether each reference event has at least
one detection within a \(\pm 1\)-step window around its
phase boundary. All three detectors achieve
\(\mathrm{event\_recall@1}=1.0\), showing that every reference
event is detected within this window.
Together, these results support using the selected features
for online keyframe detection and motivate their use for
visual-memory updates in RePlanMem-VLA. The experimental protocol and extended results are provided
in Appendix~\ref{app:rq2-vla}.

 \vspace{-0.5em}
\subsection{RQ3: Surprise-gated visual-memory updates}
\label{sec:rq3}
\vspace{-0.5em}

We finally investigate whether plan disagreement can guide visual-memory updates in tasks that require information from earlier observations.
Specifically, taking the \texttt{RememberColor} task as an example, we examine whether the selected observations are written and remain
available in memory until they are needed for action selection. 


\subsubsection{Direct surprise-gated writing}
\vspace{-0.5em}

We first evaluate the simplest memory-update rule, in which each
detected event directly writes the immediately preceding top-view
observation to the FIFO.
We examine the resulting memory-write behavior using
$\pi_{0.5}$ and OpenVLA-OFT on \texttt{RememberColor3-VLA-v0}, testing different detector thresholds for writing keyframes to memory.
At the 95th-percentile threshold, the cue observation is written
in fewer than \(20\%\) of the evaluated $\pi_{0.5}$ episodes
and in none of the evaluated OpenVLA-OFT episodes, whereas
the runs with an 80th-percentile threshold achieve cue-writing
rates of \(48\%\) and \(85\%\), respectively.
Disagreement peaks also often span several replanning steps,
leading to repeated writes of observations associated with the
same keyframe event (Fig.~\ref{fig:rq3-writes}).
These observations motivate using a lower detection threshold
and selecting visually distinct frames rather than writing
an observation at every query above the threshold.


\subsubsection{Redundancy-aware memory selection}
\vspace{-0.5em}

We next replace direct writes with the redundancy-aware selector from
Section~\ref{sec:redundancy-aware-selection}. At each detected event, we examine a short window of preceding
observations and group visually similar frames together with the observations already stored in memory.
We then select the most recent frame from each group not represented in memory. This allows the selector to identify candidate keyframes associated with the detected plan revision while avoiding repeated writes of visually similar observations.

With the addition of clustering, cue-writing and retention rates for OpenVLA-OFT improve to \(100\%\) in the evaluated 50 episodes. However, observations of the same environment state can still be assigned to different clusters because of changes in robot pose,
leading to frequent memory overwrites. With a bounded FIFO memory, these redundant writes can evict task-relevant information observed during the early stages
of the episode. This motivates masking the robot during visual comparison to suppress redundant updates caused by manipulator motion.

\subsubsection{Robot-masked redundancy filtering}
\vspace{-0.5em}
We next add robot masking to the frame-selection procedure and evaluate its effect on memory updates.
The masks shown in Figure~\ref{fig:rq3-masks} are applied only during visual comparison, while the original RGB observations are stored in memory and supplied to the VLA. 
For OpenVLA-OFT, across 50 evaluated episodes, robot masking
nearly halves the number of memory writes from 336 to 169
and reduces FIFO evictions by almost a factor of ten,
from 186 to 19.
At the same time, the task-relevant keyframe is written and retained until the action phase in all \(50/50\) episodes.

We thus obtain a mechanism that uses plan disagreement to successfully detect keyframes and retain them in visual memory.

 \vspace{-0.5em}
\subsubsection{Transfer to real-robot replay}
\vspace{-0.5em}

Finally, we evaluate the same event-detection and redundancy-filtering
pipeline through offline replay
of recorded real-robot trajectories, testing the event detector,
visual clustering, and robot-masked similarity outside simulation.

\begin{figure*}[t]
\centering
\includegraphics[width=\textwidth]{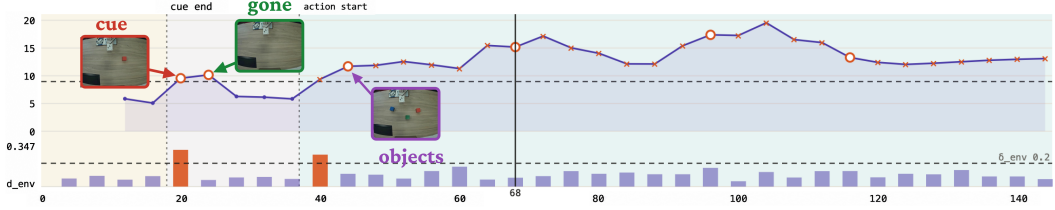}
\caption{
\textbf{Real-robot replay of the surprise-based event detector and
redundancy-aware memory-selection pipeline.} White highlights indicate memory updates, while red crosses mark
detector activations for which the redundancy filter prevents
overwriting existing memory frames.
}
\label{fig:rq3-real-data}
\vspace{-0.5em}
\end{figure*}

Figure~\ref{fig:rq3-real-data} illustrates how the detector responds
to keyframes associated with phase transitions in the
\texttt{RememberColor} task, while robot masking suppresses redundant memory overwrites during the action phase.


\vspace{-1em}

\section{Limitations and Future Work}
\label{sec:limitations}
 \vspace{-0.5em}

Our experiments show that disagreement between overlapping action
chunks can accurately identify task-relevant keyframes. However, we do
not claim that such disagreement is specific to semantic events.
Substantial plan revisions may also arise for other reasons, including
policy prediction errors. The proposed signal should therefore be
interpreted as an indicator of potentially important change rather than
as an unambiguous semantic event detector.

Another limitation concerns the temporal structure of the evaluated
tasks. The task phases in our experiments have relatively clear
boundaries. With an execution stride of \(H=4\), short
phases may occur entirely between two replanning queries and therefore
be missed, whereas gradual transitions may distribute the corresponding
plan revision over several queries and produce no pronounced peak.
How disagreement behaves under less distinct phase transitions remains
to be investigated.

Our current implementation stores selected observations as full images
in a small visual memory and uses visual clustering to suppress
redundant entries. A promising direction is therefore to develop more compact
representations of stored observations, together with comparison and
aggregation mechanisms that preserve task-relevant information while
making historical context easier for the policy to use.

Finally, the current surprise estimate is defined relative to
task-conditioned statistics accumulated from previous replanning
queries, and the write threshold is selected empirically. Consequently,
the present formulation requires calibration data for each task and has
not yet been shown to transfer directly to previously unseen tasks
without accumulating new reference statistics and selecting an
appropriate threshold.

\section{Conclusion}
\label{sec:conclusion}
 \vspace{-0.5em}

We investigated whether revisions of overlapping action chunks can
serve as an internal signal of observation importance in long-horizon
robot control. Across single-task  autoregressive and Flow-Matching
policies, we find that disagreement between consecutive plans contains
a consistent signal of semantic task transitions, with complementary
distributional, cumulative, and causal temporal measures capturing
different aspects of plan revision. We further show that this signal
transfers to pretrained OpenVLA-OFT and $\pi_{0.5}$ policies across
different replanning frequencies and remains predictive even when
control performance varies substantially.
Building on these observations, we use disagreement-derived surprise to
drive selective visual-memory updates in a memory-augmented VLA policy.
The resulting framework connects three stages that are usually treated
separately: replanning, event detection, and memory formation. Rather
than introducing an external keyframe selector, the policy reuses a signal that naturally arises from its own action predictions to decide
when the observation history should be updated.
Overall, our results suggest that changes in a policy's intended future behavior provide a useful source of temporal proposals for sparse memory updates in partially observable, multi-stage manipulation. 



{
\small
\bibliographystyle{plainnat}
\bibliography{references}

@article{pi05,
  title={$\pi_{0.5}$: a Vision-Language-Action Model with Open-World Generalization},
  author={{Physical Intelligence} and Black, Kevin and Brown, Noah and
          Darpinian, James and Dhabalia, Karan and Driess, Danny and
          Esmail, Adnan and Equi, Michael and Finn, Chelsea and
          Fusai, Niccolo and Galliker, Manuel Y. and Ghosh, Dibya and
          Groom, Lachy and Hausman, Karol and Ichter, Brian and
          Jakubczak, Szymon and Jones, Tim and Ke, Liyiming and
          LeBlanc, Devin and Levine, Sergey and Li-Bell, Adrian and
          Mothukuri, Mohith and Nair, Suraj and Pertsch, Karl and
          Ren, Allen Z. and Shi, Lucy Xiaoyang and Smith, Laura and
          Springenberg, Jost Tobias and Stachowicz, Kyle and Tanner, James and
          Vuong, Quan and Walke, Homer and Walling, Anna and Wang, Haohuan and
          Yu, Lili and Zhilinsky, Ury},
  journal={arXiv preprint arXiv:2504.16054},
  year={2025},
  url={https://arxiv.org/abs/2504.16054}
}

@article{cherepanov2026muvla,
  title={{\textmu}VLA: On Recurrent Memory for Partially Observable Manipulation in {VLA} Models},
  author={Cherepanov, Egor and Kachaev, Nikita and Zelezetsky, Daniil and Bulatov, Aydar and
          Pshenitsyn, Artem and Kuratov, Yuri and Skrynnik, Alexey and Panov, Aleksandr I. and
          Kovalev, Alexey K.},
  journal={arXiv preprint arXiv:2606.12497},
  year={2026},
  url={https://arxiv.org/abs/2606.12497}
}

@article{chen2025keyframechaining,
  title={Non-Markovian Long-Horizon Robot Manipulation via Keyframe Chaining},
  author={Chen, Yipeng and Tan, Wentao and Zhu, Lei and Li, Fengling and Li, Jingjing and Yang, Guoli and Shen, Heng Tao},
  journal={arXiv preprint arXiv:2603.01465},
  year={2026},
  url={https://arxiv.org/abs/2603.01465}
}

@inproceedings{chung2025objectcentric,
  title={Rethinking Progression of Memory State in Robotic Manipulation: An Object-Centric Perspective},
  author={Chung, Nhat and Hanyu, Taisei and Nguyen, Toan and Le, Huy and
          Bumgarner, Frederick and Nguyen, Duy Minh Ho and Vo, Khoa and
          Yamazaki, Kashu and Rainwater, Chase and Kieu, Tung and
          Nguyen, Anh and Le, Ngan},
  booktitle={Proceedings of the AAAI Conference on Artificial Intelligence},
  volume={40},
  number={5},
  pages={3407--3415},
  year={2026},
  doi={10.1609/aaai.v40i5.37337}
}

@article{guo2025chameleon,
  title={Chameleon: Control-Indexed Prospective Memory for Visuomotor Manipulation},
  author={Guo, Xinying and Jiang, Chenxi and Kim, Hyun Bin and Han, Yuhang and Sun, Ying and Xiao, Yang and Yang, Jianfei},
  journal={arXiv preprint arXiv:2603.24576},
  year={2026},
  url={https://arxiv.org/abs/2603.24576}
}

@inproceedings{haresh2026notestoself,
  title={Notes-to-Self: Scratchpad Augmented VLAs for Memory Dependent Manipulation Tasks},
  author={Haresh, Sanjay and Dijkman, Daniel and Bhattacharyya, Apratim and Memisevic, Roland},
  booktitle={IEEE International Conference on Robotics and Automation (ICRA)},
  year={2026},
  url={https://arxiv.org/abs/2602.21013}
}

@inproceedings{jing2025mixturehorizons,
  title={Mixture of Horizons in Action Chunking},
  author={Jing, Dong and Wang, Gang and Liu, Jiaqi and Tang, Weiliang and Sun, Zelong and Yao, Yunchao and Wei, Zhenyu and Liu, Yunhui and Lu, Zhiwu and Ding, Mingyu},
  booktitle={Proceedings of the 43rd International Conference on Machine Learning},
  series={Proceedings of Machine Learning Research},
  volume={306},
  year={2026}
}

@article{li2026rememvla,
  title={ReMem-VLA: Empowering Vision-Language-Action Model with Memory via Dual-Level Recurrent Queries},
  author={Li, Hang and Shen, Fengyi and Chen, Dong and Yang, Liudi and
          Wang, Xudong and Shi, Jinkui and Bing, Zhenshan and Liu, Ziyuan and
          Knoll, Alois},
  journal={arXiv preprint arXiv:2603.12942},
  year={2026},
  url={https://arxiv.org/abs/2603.12942}
}

@article{li2026spatialmem,
  title={Spatial Memory for Out-of-Vision Manipulation in Vision-Language-Action},
  author={Li, Pengteng and Guo, Weiyu and Zhang, He and Cai, Tiefu and
          He, Xiao and Guo, Yandong and Xiong, Hui},
  journal={arXiv preprint arXiv:2605.22283},
  year={2026},
  url={https://arxiv.org/abs/2605.22283}
}

@inproceedings{mark2025bpp,
  title={Long-Context Robot Imitation Learning by Focusing on Key History Frames},
  author={Mark, Max Sobol and Attarian, Maria and Fu, Chuyuan and Dwibedi, Debidatta and Liang, Jacky and Shah, Dhruv and Kumar, Aviral},
  booktitle={Proceedings of Robotics: Science and Systems},
  year={2026},
  address={Sydney, Australia},
  month={July},
  doi={10.15607/RSS.2026.XXII.201}
}

@inproceedings{cherepanov2026memory,
  title     = {Memory, Benchmark \& Robots: A Benchmark for Solving Complex Tasks with Reinforcement Learning},
  author    = {Egor Cherepanov and Nikita Kachaev and Alexey Kovalev and Aleksandr I. Panov},
  booktitle = {The Fourteenth International Conference on Learning Representations},
  year      = {2026},
  url       = {https://openreview.net/forum?id=9cLPurIZMj}
}

@misc{cherepanov2026mikasarobovlabenchmarkingmemoryvla,
      title={MIKASA-Robo-VLA: Benchmarking Memory in VLA Models for Long-Horizon Manipulation}, 
      author={Egor Cherepanov and Nikita Kachaev and Aleksandr I. Panov and Alexey K. Kovalev},
      year={2026},
      eprint={2610.00604},
      archivePrefix={arXiv},
      primaryClass={cs.LG},
      url={https://arxiv.org/abs/2610.00604}, 
}

@inproceedings{james2022qattention,
  title     = {Coarse-To-Fine {Q}-Attention: Efficient Learning for Visual Robotic Manipulation via Discretisation},
  author    = {James, Stephen and Wada, Kentaro and Laidlow, Tristan and Davison, Andrew J.},
  booktitle = {Proceedings of the IEEE/CVF Conference on Computer Vision and Pattern Recognition (CVPR)},
  pages     = {13739--13748},
  year      = {2022}
}

@inproceedings{zhao2023act,
  title     = {Learning Fine-Grained Bimanual Manipulation with Low-Cost Hardware},
  author    = {Zhao, Tony Z. and Kumar, Vikash and Levine, Sergey and Finn, Chelsea},
  booktitle = {Robotics: Science and Systems},
  year      = {2023}
}

@inproceedings{chi2023diffusionpolicy,
  title     = {Diffusion Policy: Visuomotor Policy Learning via Action Diffusion},
  author    = {Chi, Cheng and Feng, Siyuan and Du, Yilun and Xu, Zhenjia and Cousineau, Eric
               and Burchfiel, Benjamin and Song, Shuran},
  booktitle = {Robotics: Science and Systems},
  year      = {2023}
}

@inproceedings{lipman2023flow,
  title     = {Flow Matching for Generative Modeling},
  author    = {Lipman, Yaron and Chen, Ricky T. Q. and Ben-Hamu, Heli
               and Nickel, Maximilian and Le, Matt},
  booktitle = {International Conference on Learning Representations},
  year      = {2023}
}

@article{kim2025oft,
  title   = {Fine-Tuning Vision-Language-Action Models: Optimizing Speed and Success},
  author  = {Kim, Moo Jin and Finn, Chelsea and Liang, Percy},
  journal = {arXiv preprint arXiv:2502.19645},
  year    = {2025},
  url     = {https://arxiv.org/abs/2502.19645}
}

@inproceedings{liu2024bid,
  title     = {Bidirectional Decoding: Improving Action Chunking via Guided Test-Time Sampling},
  author    = {Liu, Yuejiang and Ibn Hamid, Jubayer and Xie, Annie and
               Lee, Yoonho and Du, Maximilian and Finn, Chelsea},
  booktitle = {International Conference on Learning Representations},
  year      = {2025}
}

@inproceedings{shi2026memoryvla,
  title   = {{MemoryVLA}: Perceptual-Cognitive Memory in Vision-Language-Action Models for Robotic Manipulation},
  author  = {Shi, Hao and Xie, Bin and Liu, Yingfei and Sun, Lin and Liu, Fengrong and Wang, Tiancai and Zhou, Erjin and Fan, Haoqiang and Zhang, Xiangyu and Huang, Gao},
  booktitle = {The Fourteenth International Conference on Learning Representations},
  year    = {2026},
  url     = {https://openreview.net/forum?id=54U3XHf7qq}
}

@article{chen2026rmbench,
  title   = {{RMBench}: Memory-Dependent Robotic Manipulation Benchmark with Insights into Policy Design},
  author  = {Chen, Tianxing and Wang, Yuran and Li, Mingleyang and Qin, Yan and
             Shi, Hao and Li, Zixuan and Hu, Yifan and Zhang, Yingsheng and
             Wang, Kaixuan and Chen, Yue and Wang, Hongcheng and Wang, Junjie and
             Yang, Tianhang and Xu, Renjing and Wu, Ruihai and Mu, Yao and
             Yang, Yaodong and Dong, Hao and Luo, Ping},
  journal = {arXiv preprint arXiv:2603.01229},
  year    = {2026},
  url     = {https://arxiv.org/abs/2603.01229}
}

@article{jang2025contextvla,
  title   = {{ContextVLA}: Vision-Language-Action Model with Amortized Multi-Frame Context},
  author  = {Jang, Huiwon and Yu, Sihyun and Kwon, Heeseung and Jeon, Hojin and Seo, Younggyo and Shin, Jinwoo},
  journal = {arXiv preprint arXiv:2510.04246},
  year    = {2025},
  url     = {https://arxiv.org/abs/2510.04246}
}

@article{lei2026robomemarena,
  title   = {{RoboMemArena}: A Comprehensive and Challenging Robotic Memory Benchmark},
  author  = {Lei, Huashuo and Song, Wenxuan and Zhang, Huarui and Pei, Jieyuan and Chen, Jiayi and Yan, Haodong and Zhao, Han and Ding, Pengxiang and Zhang, Zhipeng and Huang, Lida and Wang, Donglin and Wang, Yan and Li, Haoang},
  journal = {arXiv preprint arXiv:2605.10921},
  year    = {2026},
  url     = {https://arxiv.org/abs/2605.10921}
}

@inproceedings{sridhar2025memer,
  title={{MemER}: Scaling Up Memory for Robot Control via Experience Retrieval},
  author={Sridhar, Ajay and Pan, Jennifer and Sharma, Satvik and Finn, Chelsea},
  booktitle={The Fourteenth International Conference on Learning Representations},
  year={2026},
  url={https://arxiv.org/abs/2510.20328}
}

@inproceedings{liu2026palm,
  title   = {{PALM}: Progress-Aware Policy Learning via Affordance Reasoning for Long-Horizon Robotic Manipulation},
  author  = {Liu, Yuanzhe and Zhu, Jingyuan and Mo, Yuchen and Li, Gen and Cao, Xu and Jin, Jin and Shen, Yifan and Li, Zhengyuan and Yu, Tianjiao and Yuan, Wenzhen and Ding, Fangqiang and Lourentzou, Ismini},
  booktitle = {Proceedings of the IEEE/CVF Conference on Computer Vision and Pattern Recognition (CVPR)},
  pages   = {28096--28110},
  year    = {2026}
}

@inproceedings{buamanee2026bihil,
  title   = {{Bi-HIL}: Bilateral Control-Based Multimodal Hierarchical Imitation Learning via Subtask-Level Progress Rate and Keyframe Memory for Long-Horizon Contact-Rich Robotic Manipulation},
  author  = {Buamanee, Thanpimon and Kobayashi, Masato and Uranishi, Yuki},
  booktitle = {IEEE/RSJ International Conference on Intelligent Robots and Systems (IROS)},
  year    = {2026},
  url     = {https://arxiv.org/abs/2603.13315}
}

@article{feng2026procvlm,
  title   = {{ProcVLM}: Learning Procedure-Grounded Progress Rewards for Robotic Manipulation},
  author  = {Feng, Youhe and Shi, Hansen and Li, Haoyang and Guo, Xinlei and Wang, Yang and Zhang, Chengyang and Zhang, Jinkai and Zhang, Xiaohan and Tang, Jie and Zhang, Jing},
  journal = {arXiv preprint arXiv:2605.08774},
  year    = {2026},
  url     = {https://arxiv.org/abs/2605.08774}
}

@article{qiu2026efficientlonghorizonvla,
  title   = {Efficient Long-Horizon Vision-Language-Action Models via Static-Dynamic Disentanglement},
  author  = {Qiu, Weikang and Lei, Huashuo and Huang, Tinglin and Ying, Rex},
  journal = {arXiv preprint arXiv:2602.03983},
  year    = {2026},
  url     = {https://arxiv.org/abs/2602.03983}
}

@article{wang2026nativemem,
  title   = {{NativeMEM}: Native Memory Compression for Long-Horizon Robotic Manipulation},
  author  = {Wang, Ziye and Shi, Modi and Ni, Chaojun and Yang, Jiazhi and Li, Mengdi and Su, Zhizhong and Lin, Tianwei and Li, Hongyang},
  journal = {arXiv preprint arXiv:2607.06678},
  year    = {2026}
}

@inproceedings{shi2023waypoint,
  title   = {Waypoint-Based Imitation Learning for Robotic Manipulation},
  author  = {Shi, Lucy Xiaoyang and Sharma, Archit and Zhao, Tony Z. and Finn, Chelsea},
  booktitle = {Proceedings of the 7th Conference on Robot Learning},
  series  = {Proceedings of Machine Learning Research},
  volume  = {229},
  pages   = {2195--2209},
  publisher = {PMLR},
  year    = {2023},
  url     = {https://proceedings.mlr.press/v229/shi23b.html}
}

@article{zeng2026kemo,
  title   = {{KEMO}: Event-Driven Keyframe Memory for Long-Horizon Robot Manipulation with VLA Policies},
  author  = {Zeng, Yihan and Ye, Minghao and Chen, Yiyuan and Shentu, Yide and Wu, Philipp and Yan, Zike and Li, Zhongyu},
  journal = {arXiv preprint arXiv:2606.23589},
  year    = {2026},
  url     = {https://arxiv.org/abs/2606.23589}
}

@article{zeng2026revisitingopenloop,
  title   = {Revisiting Open-Loop Execution in Robotics: Toward Reactive, Higher-Performing Policies},
  author  = {Zeng, Michael and Agarwal, Abhinav and Bati, Ajay and Lee, Brian and Ancha, Siddharth and Tedrake, Russ},
  journal = {arXiv preprint arXiv:2608.15938},
  year    = {2026},
  url     = {https://arxiv.org/abs/2608.15938}
}

@article{chen2026dreamchunk,
  title   = {{DREAM-Chunk}: Reactive Action Chunking with Latent World Model},
  author  = {Chen, Wenxi and Zhang, Kaidi and Lin, Chi and Zhang, Zhiyuan and She, Yu and Liu, Yuejiang and Yeh, Raymond A. and Mou, Shaoshuai and Gu, Yan},
  journal = {arXiv preprint arXiv:2606.18589},
  year    = {2026},
  url     = {https://arxiv.org/abs/2606.18589}
}

@article{lazzati2026doesactionchunkingimprove,
  title   = {Why Does Action Chunking Improve Behavioral Cloning Performance in Robotic Control?},
  author  = {Lazzati, Filippo and Stachowicz, Kyle and Chen, William and Metelli, Alberto Maria and Wagenmaker, Andrew and Levine, Sergey},
  journal = {arXiv preprint arXiv:2608.02547},
  year    = {2026},
  url     = {https://arxiv.org/abs/2608.02547}
}

@article{yi2026criticlooptrisystemvla,
  title   = {Critic in the Loop: A Tri-System VLA Framework for Robust Long-Horizon Manipulation},
  author  = {Yi, Pengfei and Ma, Yingjie and Xu, Wenjiang and Hao, Yanan and Gan, Shuai and Li, Wanting and Zhong, Shanlin},
  journal = {arXiv preprint arXiv:2603.05185},
  year    = {2026},
  url     = {https://arxiv.org/abs/2603.05185}
}

@inproceedings{chen2016xgboost,
  author    = {Tianqi Chen and Carlos Guestrin},
  title     = {{XGBoost}: A Scalable Tree Boosting System},
  booktitle = {Proceedings of the 22nd ACM SIGKDD International
               Conference on Knowledge Discovery and Data Mining},
  year      = {2016},
  pages     = {785--794},
  doi       = {10.1145/2939672.2939785}
}

@inproceedings{prokhorenkova2018catboost,
  author    = {Liudmila Prokhorenkova and Gleb Gusev and
               Aleksandr Vorobev and Anna Veronika Dorogush and
               Andrey Gulin},
  title     = {{CatBoost}: Unbiased Boosting with Categorical Features},
  booktitle = {Advances in Neural Information Processing Systems},
  volume    = {31},
  year      = {2018},
  url       = {https://proceedings.neurips.cc/paper/2018/hash/14491b756b3a51daac41c24863285549-Abstract.html}
}

@article{hochreiter1997lstm,
  author  = {Sepp Hochreiter and J{\"u}rgen Schmidhuber},
  title   = {Long Short-Term Memory},
  journal = {Neural Computation},
  volume  = {9},
  number  = {8},
  pages   = {1735--1780},
  year    = {1997},
  doi     = {10.1162/neco.1997.9.8.1735}
}
}

\clearpage
\appendix

\part{\vspace{-30pt}}
\vspace{-40pt}
\parttoc


\section{Additional Analysis for RQ1}
\label{app:rq1-analysis}
\vspace{-0.5em}
This appendix details the experimental setup, evaluation protocol,
and supporting analyses for Section~\ref{sec:rq1-metric-screening}.

\vspace{-0.5em}
\subsection{Experimental setup and reference events}
\label{app:rq1-setup}

\paragraph{Tasks and training.}
The single-task screening uses eight MIKASA-Robo-VLA
tasks~\citep{cherepanov2026mikasarobovlabenchmarkingmemoryvla}:
\texttt{RememberColor3-VLA-v0},
\texttt{ShellGameColorLampTouch-VLA-v0},
\texttt{ShellGameShuffleColorLampTouch-VLA-v0},
\texttt{ShellGameShuffleTouch-VLA-v0},
\texttt{FindImposterShape3-VLA-v0},
\texttt{InterceptSlow-VLA-v0},
\texttt{RotateLenientPos-VLA-v0},
and \texttt{TakeItBack-VLA-v0}.
Each policy is trained from scratch on 250 demonstrations
of a single task.
A separate policy is trained for each combination of task,
architecture, and context regime for 150 epochs,
using training seed 42.

\begin{wraptable}{r}{0.52\textwidth}
\vspace{-1.5em}
\centering
\footnotesize
\setlength{\tabcolsep}{4pt}
\renewcommand{\arraystretch}{1.1}
\caption{\textbf{Architecture and training settings
for the single-task policies in RQ1.}}
\label{tab:rq1-policy-configuration}
\begin{tabularx}{\linewidth}{
@{}>{\raggedright\arraybackslash}p{0.43\linewidth}Y@{}
}
\toprule
\textbf{Setting} & \textbf{Configuration} \\
\midrule
\multicolumn{2}{@{}l}{\textbf{Shared settings}} \\
\addlinespace[2pt]

Visual encoder
& NatureCNN, 512-D \\

Proprioception encoder
& MLP \(7\rightarrow128\rightarrow128\) \\

Context Transformer
& 4 layers, 8 attention heads \\

Hidden / FFN dimension
& 256 / 1024 \\

Dropout
& 0.1 \\

Optimizer
& AdamW \\

Learning rate
& \(3\times10^{-4}\), cosine decay to \(10^{-6}\) \\

Weight decay
& \(10^{-4}\) \\

Batch size
& 4 episodes \\

Gradient clipping
& 1.0 \\

\midrule
\multicolumn{2}{@{}l}{\textbf{Architecture-specific settings}} \\
\addlinespace[2pt]

Training objective
& Masked action MSE (autoregressive)
\newline
Masked velocity MSE (flow matching) \\

Flow action head
& 2 Transformer layers \\

Flow integration
& Heun, 10 steps \\

\bottomrule
\end{tabularx}
\vspace{-3.5em}
\end{wraptable}

\paragraph{Observations and policy architectures.}
Each observation contains top-view and wrist-camera RGB
images at \(128\times128\) resolution and a seven-dimensional
proprioceptive state describing the end-effector pose and
gripper position.
The images are scaled to \([0,1]\), concatenated along the
channel dimension, and encoded by NatureCNN.
Proprioception and action targets are standardized using
training-set statistics.
For \texttt{RotateLenientPos-VLA-v0}, the policy additionally
receives the numerical target angle from the task instruction.

Both architectures use the same visual and proprioceptive
encoder configurations and context Transformer settings.
Their shared settings and architecture-specific components
are summarized in Table~\ref{tab:rq1-policy-configuration}.

\paragraph{Execution and evaluation.}
Both architectures predict action chunks of length \(K=8\)
and execute one action before replanning (\(H=1\)).
During Flow-Matching evaluation, a fixed Gaussian noise
tensor is reused across queries within each rollout to
prevent stochastic resampling from introducing artificial
plan disagreement.
For each task and context regime, each policy is evaluated
on 50 closed-loop episodes.

\paragraph{Reference keyframes.}
Task-specific simulator events provide the reference
keyframe labels.
For a phase transition between observations \(b-1\) and
\(b\), we define the reference event window as
\[
    \mathcal I_e=[b-1,b].
\]
Both the final observation of the preceding phase and the
first observation of the following phase are labelled
as keyframes.
All remaining observations are labelled as non-keyframes.
The task-specific reference events are listed in
Table~\ref{tab:rq1-reference-events}.

\vspace{-0.5em}
\subsection{Task-specific results }
\label{app:rq1-rankings}
\vspace{-0.5em}

Table~\ref{tab:rq1-task-best} reports the best-performing
disagreement metric and its Frame ROC-AUC for each task,
architecture, and context regime.

\begin{table}[t]
\centering
\footnotesize
\setlength{\tabcolsep}{4pt}
\renewcommand{\arraystretch}{1.2}
\caption{
\textbf{Best-performing disagreement metrics across the eight
RQ1 tasks.}
}
\label{tab:rq1-task-best}

\begin{tabularx}{\linewidth}{@{}lYYYY@{}}
\toprule
&
\multicolumn{2}{c}{\textbf{Autoregressive}} &
\multicolumn{2}{c}{\textbf{Flow Matching}} \\
\cmidrule(lr){2-3}
\cmidrule(lr){4-5}

\textbf{Task} &
\textbf{Full-context} &
\textbf{Current-only} &
\textbf{Full-context} &
\textbf{Current-only} \\
\midrule

\texttt{RememberColor3} &
First NMSE\newline 0.8480 &
Energy distance\newline 0.8813 &
Energy distance\newline 0.9427 &
Huber\newline 0.9173 \\

\shortstack[l]{
\texttt{ShellGameColor}\\
\texttt{LampTouch}
} &
Causal Z-score\newline 0.7980 &
Gripper disagreement\newline 0.8335 &
Gripper disagreement\newline 0.9049 &
Gripper disagreement\newline 0.8819 \\

\shortstack[l]{
\texttt{ShellGameShuffle}\\
\texttt{ColorLampTouch}
} &
EMA deviation\newline 0.5770 &
EMA deviation\newline 0.6170 &
EMA deviation\newline 0.6014 &
EMA deviation\newline 0.6174 \\

\shortstack[l]{
\texttt{ShellGameShuffle}\\
\texttt{Touch}
} &
Causal Z-score\newline 0.7063 &
Relative TV gap\newline 0.5678 &
Causal Z-score\newline 0.8051 &
EMA deviation\newline 0.7057 \\

\texttt{FindImposterShape3} &
EMA deviation\newline 0.7332 &
Energy distance\newline 0.8496 &
EMA deviation\newline 0.7916 &
Wasserstein-1\newline 0.8558 \\

\texttt{InterceptSlow} &
Translation cosine\newline 0.7396 &
Translation cosine\newline 0.7301 &
Delta-action EMA\newline 0.7283 &
Translation cosine\newline 0.7561 \\

\texttt{RotateLenientPos} &
Translation cosine\newline 0.7303 &
Translation cosine\newline 0.6908 &
CUSUM\newline 0.7650 &
CUSUM\newline 0.7949 \\

\texttt{TakeItBack} &
Delta-action EMA\newline 0.5839 &
EMA deviation\newline 0.5858 &
Wasserstein-1\newline 0.6414 &
Wasserstein-1\newline 0.6637 \\

\bottomrule
\end{tabularx}
\end{table}

\vspace{-0.5em}
\subsection{Disagreement signals on failed trajectories}
\label{app:rq1-outcomes}
\vspace{-0.5em}
\begin{wraptable}{r}{0.43\textwidth}
\vspace{-3em}
\centering
\footnotesize
\setlength{\tabcolsep}{3pt}
\renewcommand{\arraystretch}{1.1}

\caption{
\textbf{Task-specific reference events used in RQ1.}
}
\label{tab:rq1-reference-events}

\begin{tabularx}{\linewidth}{@{}lY@{}}
\toprule
\textbf{Task} & \textbf{Reference events} \\
\midrule

\texttt{RememberColor3} &
Cue disappearance and onset of manipulation. \\

\shortstack[l]{
\texttt{ShellGameColor}\\
\texttt{LampTouch}
} &
Transition from cue presentation to manipulation. \\

\shortstack[l]{
\texttt{ShellGameShuffle}\\
\texttt{ColorLampTouch}
} &
Transitions from cue presentation to shuffling
and from shuffling to manipulation. \\

\shortstack[l]{
\texttt{ShellGameShuffle}\\
\texttt{Touch}
} &
Transitions from cue presentation to shuffling
and from shuffling to manipulation. \\

\shortstack[l]{
\texttt{FindImposter}\\
\texttt{Shape3}
} &
Transitions from cue presentation to the empty phase
and from the empty phase to manipulation. \\

\texttt{InterceptSlow} &
Transition from approach to deflection,
onset of goal-directed deflection,
and the success-achieving action. \\

\shortstack[l]{
\texttt{RotateLenient}\\
\texttt{Pos}
} &
Transition from approach to rotation. \\

\texttt{TakeItBack} &
Transition from outbound motion to return motion. \\

\bottomrule
\end{tabularx}
\vspace{-7em}
\end{wraptable}

\paragraph{Execution and evaluation.}

We examine whether disagreement remains informative when the policy fails to complete the task.
Frame ROC-AUC is computed separately for successful and
failed trajectories within each task and summarized
by the median across tasks.

For autoregressive policies, median Causal Z-score
ROC-AUC on failed trajectories is \(0.700/0.691\)
under full-context/current-only conditioning,
compared with \(0.589/0.538\) on successful trajectories.
For Flow-Matching policies, Wasserstein-1 reaches
\(0.732/0.741\) on failed trajectories,
compared with \(0.670/0.640\) on successful trajectories.
The direction of the difference depends on the measure,
with Flow-Matching EMA deviation performing better
on successful trajectories.
Overall, task failure does not lead to a systematic
decrease in disagreement-based keyframe discrimination.

\subsection{Correlation structure and metric redundancy}
\label{app:rq1-correlations}

\begin{figure*}[t]
    \centering
    \caption{
    \textbf{Correlation structure of the disagreement measures.}
    Pairwise within-episode Spearman correlations are aggregated
    across eight tasks with equal task weights.
    The top row shows autoregressive policies and the bottom row
    shows Flow-Matching policies.
    Columns correspond to full-context and current-only conditioning.
    }
    \label{fig:rq1-spearman-context}
    \includegraphics[width=\textwidth]{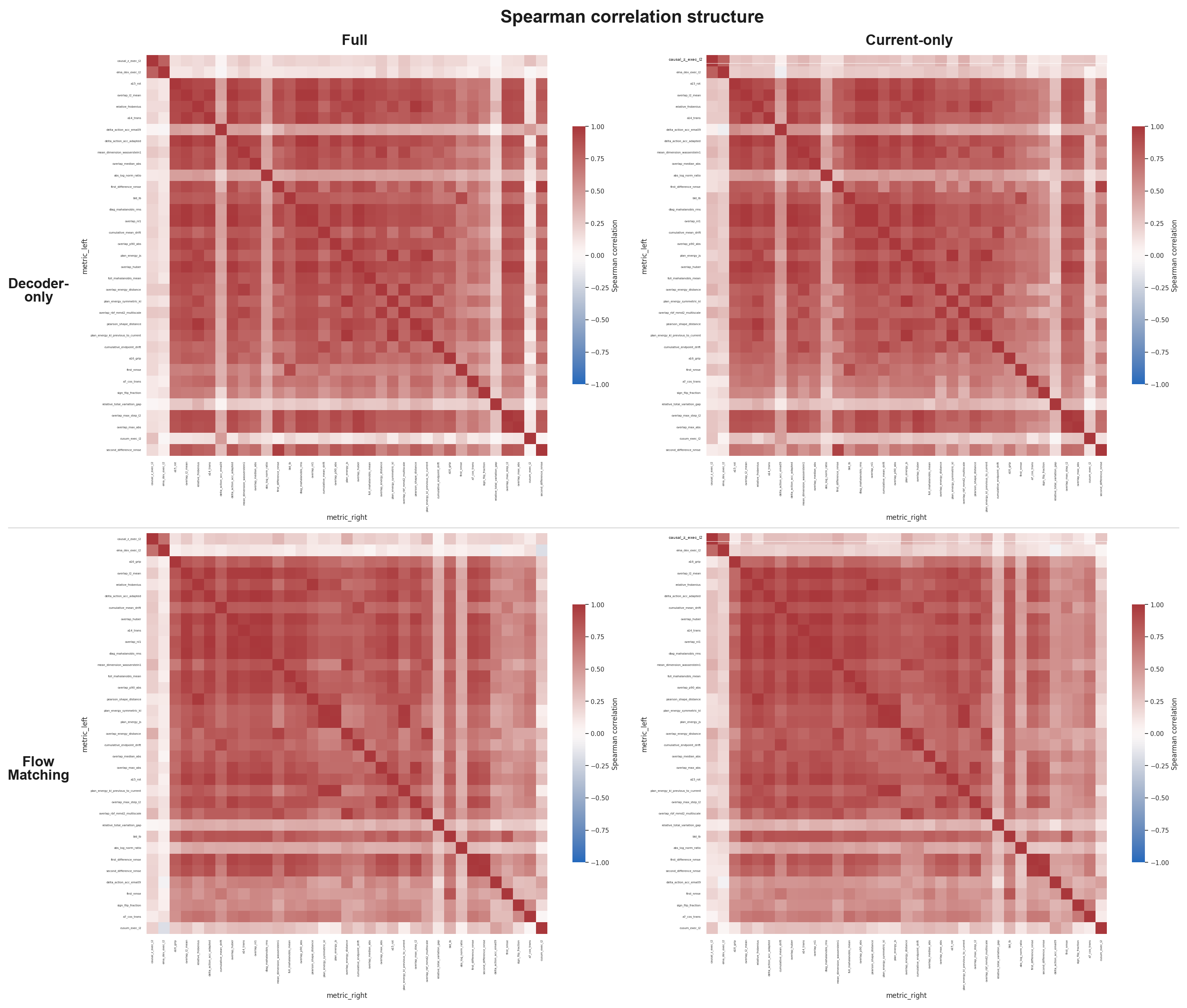}
\end{figure*}

We assess whether different disagreement measures provide
similar information by computing pairwise Spearman correlations
between their within-episode score sequences.
Correlations are summarized within each task and aggregated
with equal weight across the eight tasks.
Figure~\ref{fig:rq1-spearman-context} shows the resulting matrices
for both architectures and context regimes.

The matrices reveal substantial redundancy among magnitude-based
measures, including several point-wise, cumulative, and
distributional metrics.
Causal Z-score and EMA deviation are less strongly correlated
with many of these measures, suggesting that their temporal
normalization captures a different aspect of plan revision.

The overall correlation structure remains similar across
context regimes.
Spearman correlations between the corresponding off-diagonal
entries of the full-context and current-only matrices are
\(0.9647\) for autoregressive policies and \(0.9742\) for
Flow-Matching policies.
These analyses motivate using a compact feature set
rather than retaining many closely correlated variants.

\vspace{-1em}
\section{Additional Analysis for RQ2}
\label{app:rq2-vla}
\vspace{-0.5em}
This appendix details the experimental protocol and extended results
for Section~\ref{sec:rq2-vla-keyframe-detection}.

\vspace{-0.5em}
\subsection{Experimental setup and evaluation}
\label{app:rq2-setup}
\vspace{-0.5em}

\begingroup
\raggedright

\paragraph{Tasks and policy fine-tuning.}
Both OpenVLA-OFT and $\pi_{0.5}$ are fine-tuned on the same
19 MIKASA-Robo-VLA
tasks~\citep{cherepanov2026mikasarobovlabenchmarkingmemoryvla}:
\texttt{FindImposterColor3-VLA-v0},
\texttt{FindImposterColor5-VLA-v0},
\texttt{FindImposterColor9-VLA-v0},
\texttt{FindImposterShape3-VLA-v0},
\texttt{FindImposterShape5-VLA-v0},
\texttt{FindImposterShape9-VLA-v0},
\texttt{FindImposterShapeAndColor3x2-VLA-v0},
\texttt{FindImposterShapeAndColor3x3-VLA-v0},
\texttt{FindImposterShapeAndColor5x3-VLA-v0},
\texttt{RememberColor3-VLA-v0},
\texttt{RememberColor5-VLA-v0},
\texttt{RememberColor9-VLA-v0},
\texttt{RememberShape3-VLA-v0},
\texttt{RememberShape5-VLA-v0},
\texttt{RememberShape9-VLA-v0},
\texttt{RememberShapeAndColor3x2-VLA-v0},
\texttt{RememberShapeAndColor3x3-VLA-v0},
\texttt{RememberShapeAndColor5x3-VLA-v0},
and \texttt{TakeItBack-VLA-v0}.

\par
\endgroup

\paragraph{Detector training and evaluation.}
To train and evaluate the keyframe detectors, we collect
50 closed-loop episodes for each task, policy, and execution
regime. The resulting trajectories are divided at the episode
level into approximately 70\% training, 10\% validation,
and 20\% test subsets, keeping each episode intact and
preserving the temporal order of its observations. Detectors are trained independently for each policy
and execution regime, with decision thresholds selected
by maximizing F1 on the validation episodes.

\subsection{Within-episode detection profiles}
\label{app:rq2-qual}

\begin{figure}[t]
\centering

\begin{subfigure}[t]{0.49\linewidth}
    \centering
    \includegraphics[width=\linewidth]{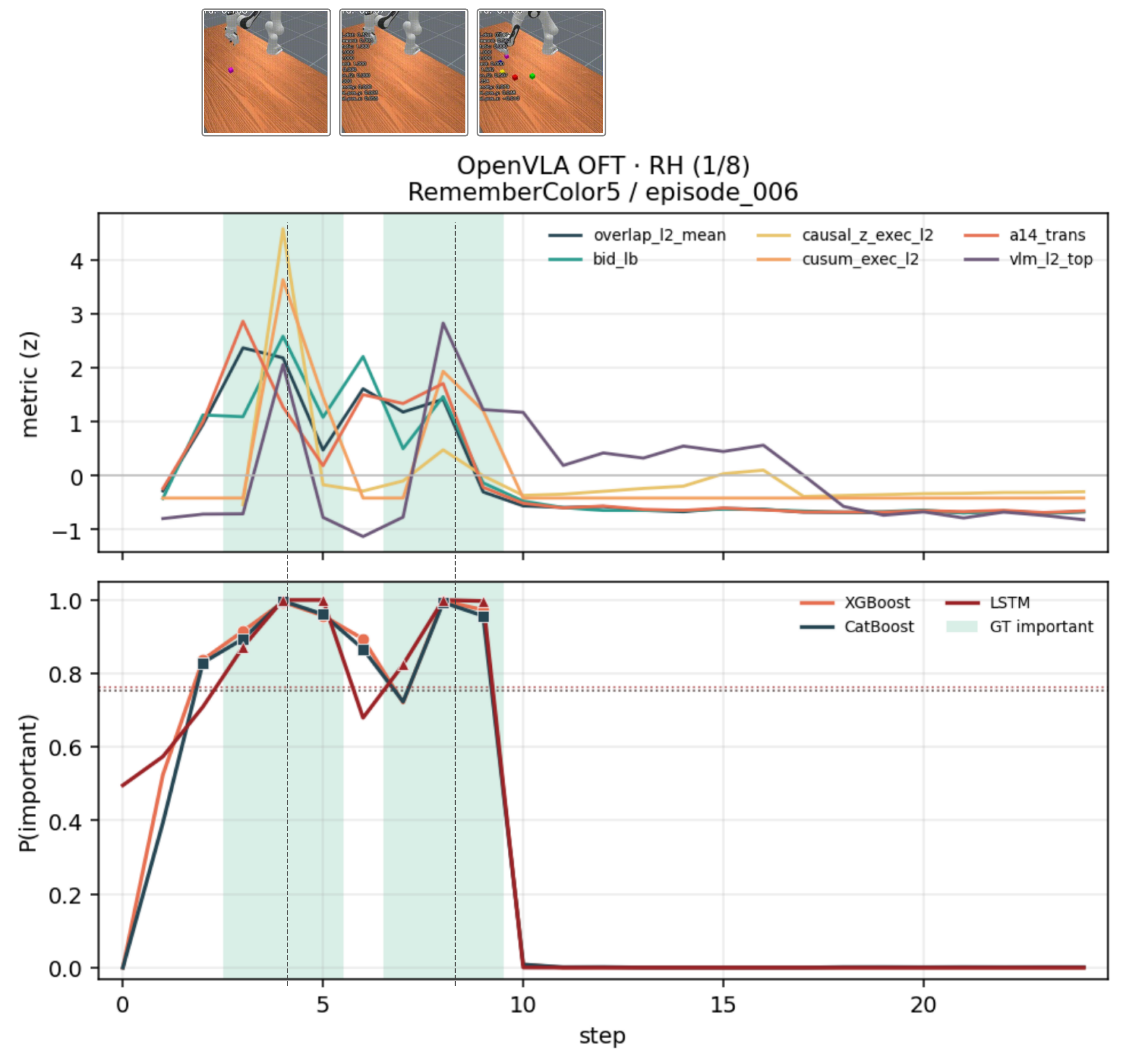}
    \caption{OpenVLA-OFT, RH1}
    \label{fig:openvla-rh1}
\end{subfigure}
\hfill
\begin{subfigure}[t]{0.49\linewidth}
    \centering
    \includegraphics[width=\linewidth]{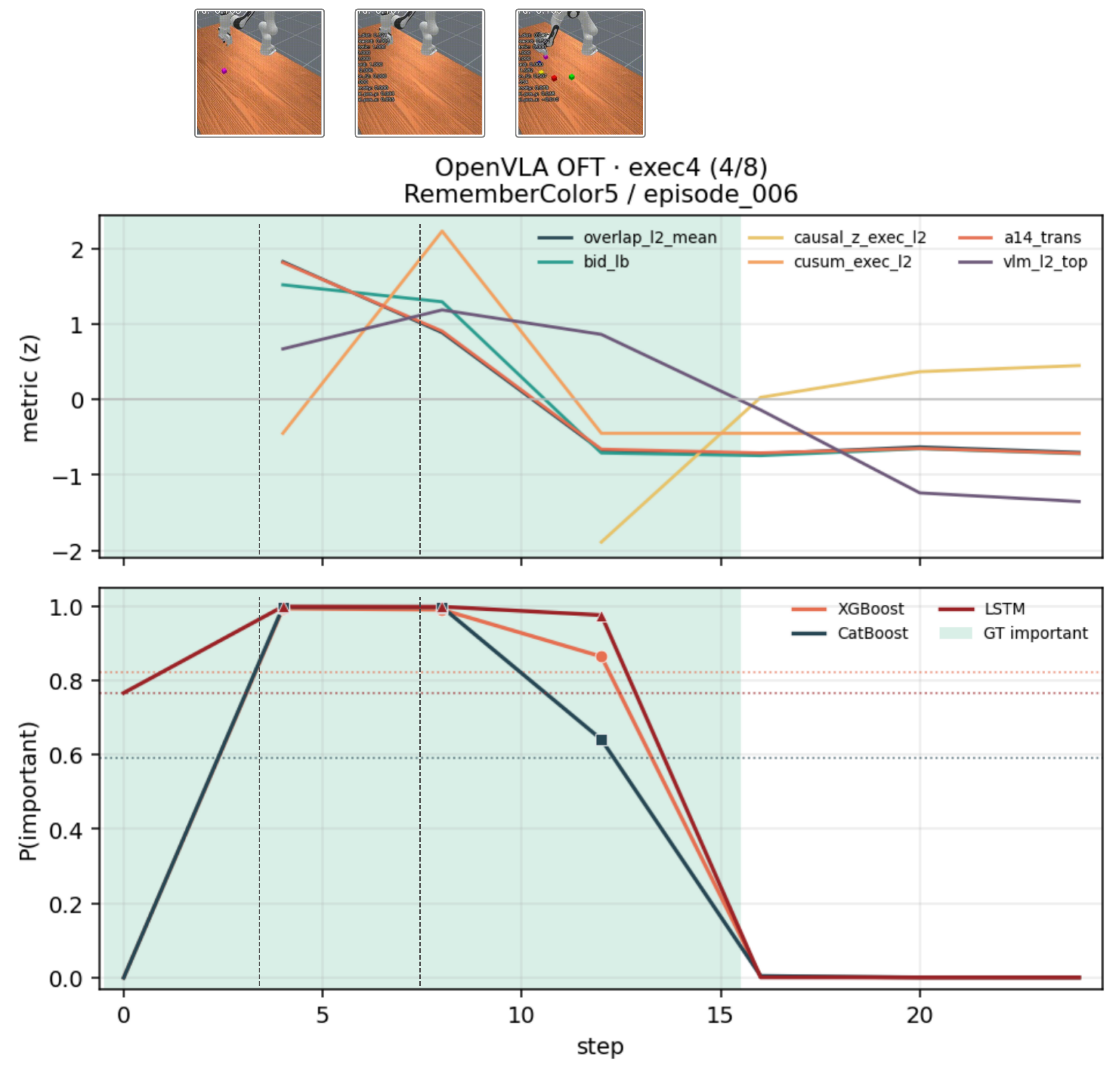}
    \caption{OpenVLA-OFT, EXEC4}
    \label{fig:openvla-exec4}
\end{subfigure}

\par\medskip

\begin{subfigure}[t]{0.49\linewidth}
    \centering
    \includegraphics[width=\linewidth]{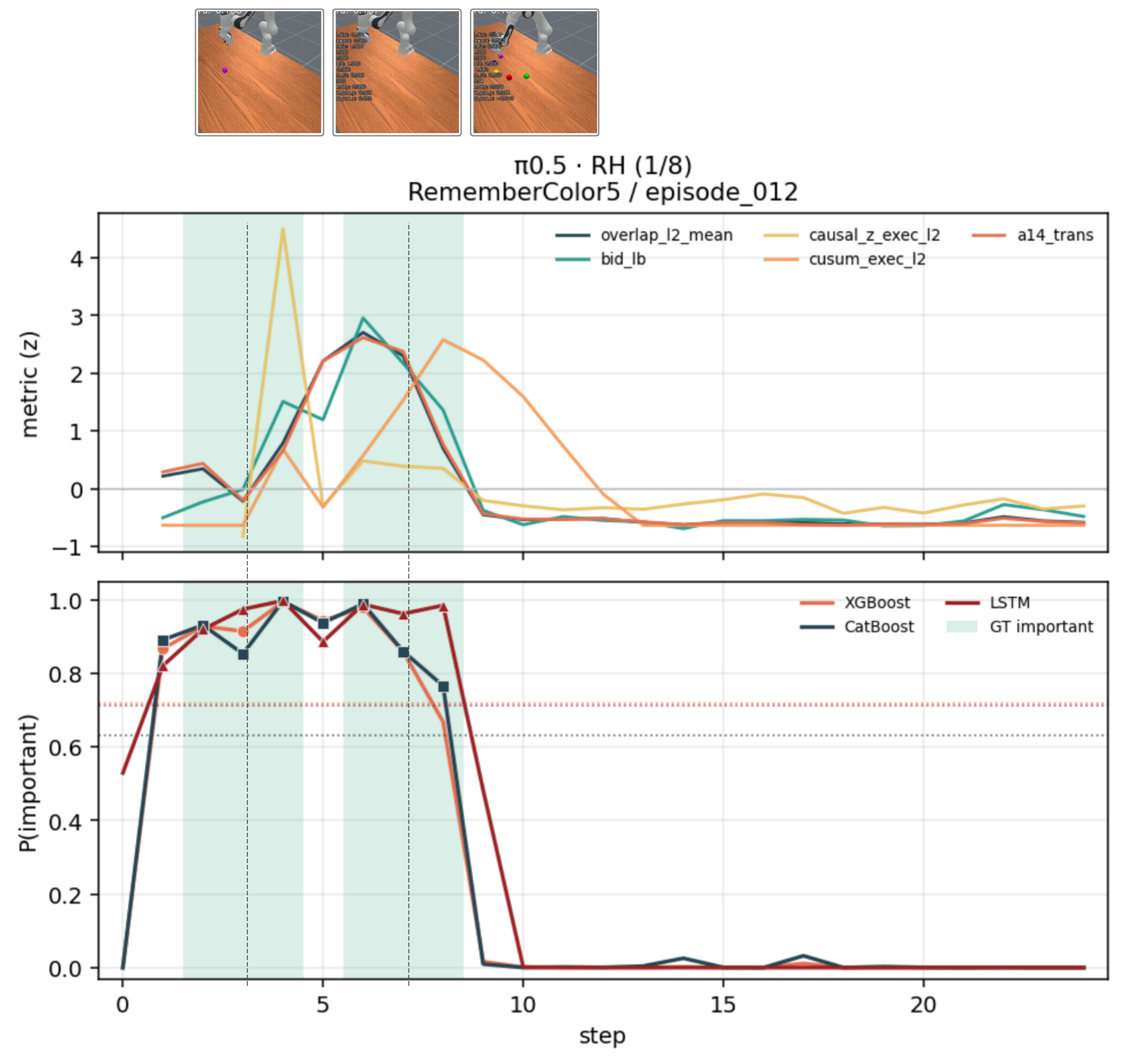}
    \caption{$\pi_{0.5}$, RH1}
    \label{fig:pi05-rh1}
\end{subfigure}
\hfill
\begin{subfigure}[t]{0.49\linewidth}
    \centering
    \includegraphics[width=\linewidth]{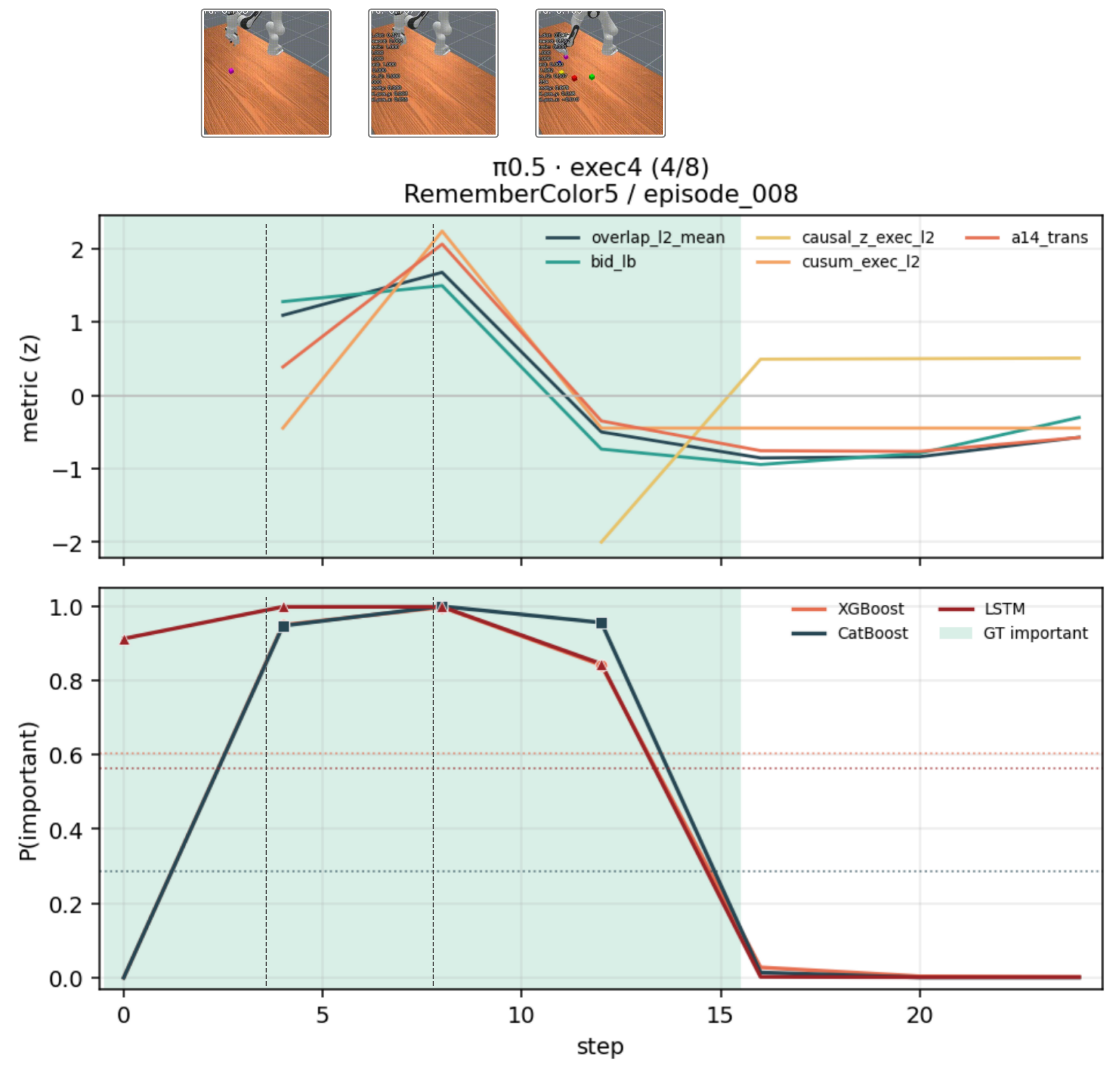}
    \caption{$\pi_{0.5}$, EXEC4}
    \label{fig:pi05-exec4}
\end{subfigure}

\caption{
\textbf{Within-episode keyframe detection on
\texttt{RememberColor5}.}
Each panel shows example observations, selected disagreement
features standardized within the episode, and keyframe
probabilities predicted by XGBoost, CatBoost, and LSTM. Horizontal dotted lines indicate detector decision thresholds.
}
\label{fig:episode-timelines}
\end{figure}

Figure~\ref{fig:episode-timelines} shows selected disagreement
features and detector outputs during
\texttt{RememberColor5} episodes for both policies under
RH1 and EXEC4.

\section{Complete Registry of Disagreement Metrics}
\label{app:all-metrics}

\providecolor{RePlanMemBlue}{HTML}{5548C8}

\providecommand{\rpmMetric}[1]{%
  {\normalfont\mdseries\color{black}\nolinkurl{#1}}%
}
\providecommand{\rpmSelectedMetric}[1]{%
  {\normalfont\mdseries\color{RePlanMemBlue}\nolinkurl{#1}}%
}

We use the aligned action sequences \(U_q,V_q\) introduced in
Eq.~\eqref{eq:chunk-overlap}, omitting the query index below.
Their length is \(L=K-H\), and each action has \(d\) components.
A tilde denotes standardization using training-action statistics:
\[
\widetilde U_{ij}
=
\frac{U_{ij}-\mu_j^{a}}{\sigma_j^{a}},
\qquad
\widetilde V_{ij}
=
\frac{V_{ij}-\mu_j^{a}}{\sigma_j^{a}}.
\]
We write
\[
E=U-V,
\qquad
\widetilde E=\widetilde U-\widetilde V.
\]
Indices are \(i=0,\ldots,L-1\) and \(j=1,\ldots,d\).

The temporal features use the first-action change
\[
s_q=\|a_{q-1,0}-a_{q,0}\|_2.
\]
We denote the mean and standard deviation of preceding values
of this signal by \(\mu_{q-1}^{s},\sigma_{q-1}^{s}\).

\begingroup
\small
\setlength{\tabcolsep}{4pt}
\renewcommand{\arraystretch}{1.35}
\setlength{\LTleft}{0pt}
\setlength{\LTright}{0pt}
\setlength{\LTcapwidth}{\textwidth}

\begin{longtable}{@{}
  >{\raggedright\arraybackslash}m{0.04\textwidth}
  >{\raggedright\arraybackslash}m{0.36\textwidth}
  >{\raggedright\arraybackslash}
  m{\dimexpr0.60\textwidth-4\tabcolsep\relax}
@{}}

\caption{\textbf{Definitions of the 35 disagreement signals.}
Colored names mark the nine action-derived features used in RQ2 and RePlanMem-VLA.}
\label{tab:all-metrics}\\

\toprule
\textbf{\#} & \textbf{Metric} & \textbf{Definition} \\
\midrule
\endfirsthead

\multicolumn{3}{@{}l}{\tablename~\thetable\ (continued)}\\[0.3em]
\toprule
\textbf{\#} & \textbf{Metric} & \textbf{Definition} \\
\midrule
\endhead

\midrule
\multicolumn{3}{r@{}}{Continued on next page}\\
\endfoot

\bottomrule
\endlastfoot

1 &
\rpmSelectedMetric{overlap_l2_mean} &
\(\displaystyle
\frac{1}{L}\sum_i\|E_i\|_2
\)
\\

2 &
\rpmSelectedMetric{bid_lb} &
\(\displaystyle
\sum_i 0.5^i\|E_i\|_2
\)
\\

3 &
\rpmSelectedMetric{causal_z_exec_l2} &
\(\displaystyle
\frac{s_q-\mu_{q-1}^{s}}
     {\sigma_{q-1}^{s}+\epsilon_t}
\)
\\

4 &
\rpmSelectedMetric{cusum_exec_l2} &
\(\displaystyle
\begin{aligned}
c_q^{\mathrm{CUSUM}}
=\max\bigl(
&0,\,
c_{q-1}^{\mathrm{CUSUM}}+s_q-\mu_0^{s}-\kappa\sigma_0^{s}
\bigr)
\end{aligned}
\)
\\

5 &
\rpmSelectedMetric{ema_dev_exec_l2} &
\(\displaystyle
\begin{aligned}
&s_q-\bar s_{q-1},\\
&\bar s_q=\alpha s_q+(1-\alpha)\bar s_{q-1}
\end{aligned}
\)
\\

6 &
\rpmSelectedMetric{a7_cos_trans} &
\(\displaystyle
\frac{1}{|\mathcal I|}
\sum_{i\in\mathcal I}c(u_i,v_i)
\)
\\

7 &
\rpmSelectedMetric{a14_trans} &
\(\displaystyle
\frac{1}{L}\sum_i
\frac{\|t_i^U-t_i^V\|_2}{\sigma_p}
\)
\\

8 &
\rpmSelectedMetric{a15_rot} &
\(\displaystyle
\frac{1}{L}\sum_i
\frac{
d_{\mathrm{SO}(3)}(\mathcal R_i^U,\mathcal R_i^V)
}{\sigma_R}
\)
\\

9 &
\rpmSelectedMetric{a16_grip} &
\(\displaystyle
\frac{1}{L}\sum_i
\frac{|g_i^U-g_i^V|}{s_g}
\)
\\

10 &
\rpmMetric{first_nmse} &
\(\displaystyle
\frac{1}{d}\sum_j\widetilde E_{0j}^{\,2}
\)
\\

11 &
\rpmMetric{overlap_nl1} &
\(\displaystyle
\frac{1}{Ld}\sum_{i,j}|\widetilde E_{ij}|
\)
\\

12 &
\rpmMetric{overlap_huber} &
\(\displaystyle
\frac{1}{Ld}\sum_{i,j}h_1(\widetilde E_{ij})
\)
\\

13 &
\rpmMetric{overlap_median_abs} &
\(\displaystyle
\operatorname{median}_{i,j}|\widetilde E_{ij}|
\)
\\

14 &
\rpmMetric{overlap_p90_abs} &
\(\displaystyle
\operatorname{Quantile}_{0.90}
\bigl(\{|\widetilde E_{ij}|\}_{i,j}\bigr)
\)
\\

15 &
\rpmMetric{overlap_max_abs} &
\(\displaystyle
\max_{i,j}|\widetilde E_{ij}|
\)
\\

16 &
\rpmMetric{overlap_max_step_l2} &
\(\displaystyle
\max_i\|\widetilde E_i\|_2
\)
\\

17 &
\rpmMetric{relative_frobenius} &
\(\displaystyle
\frac{\|\widetilde E\|_F}
{\tfrac12(\|\widetilde U\|_F+\|\widetilde V\|_F)+\epsilon}
\)
\\

18 &
\rpmMetric{abs_log_norm_ratio} &
\(\displaystyle
\left|
\log
\frac{\|\widetilde U\|_F+\epsilon}
     {\|\widetilde V\|_F+\epsilon}
\right|
\)
\\

19 &
\rpmMetric{diag_mahalanobis_rms} &
\(\displaystyle
\frac{1}{L}\sum_i
\sqrt{
\frac{1}{d}\sum_j
\left(\frac{E_{ij}}{\sigma_j^{\mathrm{cal}}}\right)^2
}
\)
\\

20 &
\rpmMetric{full_mahalanobis_mean} &
\(\displaystyle
\frac{1}{L}\sum_i
\sqrt{
E_i^\top\Sigma_a^\dagger E_i
}
\)
\\

21 &
\rpmMetric{delta_action_acc_adapted} &
\(\displaystyle
A_q^{\mathrm{ACC}}
=
\frac{1}{Ld}\sum_{i,j}\frac{|E_{ij}|}{r_j}
\)
\\

22 &
\rpmMetric{delta_action_acc_ema09} &
\(\displaystyle
\bar A_q^{\mathrm{ACC}}
=
0.9\bar A_{q-1}^{\mathrm{ACC}}
+0.1A_q^{\mathrm{ACC}}
\)
\\

23 &
\rpmMetric{pearson_shape_distance} &
\(\displaystyle
1-\operatorname{corr}
\bigl(
\operatorname{vec}\widetilde U,
\operatorname{vec}\widetilde V
\bigr)
\)
\\

24 &
\rpmMetric{plan_energy_kl_previous_to_current} &
\(\displaystyle
\operatorname{KL}(\rho\|\eta)
=
\sum_n\rho_n\log\frac{\rho_n}{\eta_n}
\)
\\

25 &
\rpmMetric{plan_energy_symmetric_kl} &
\(\displaystyle
\frac12\left[
\operatorname{KL}(\rho\|\eta)
+
\operatorname{KL}(\eta\|\rho)
\right]
\)
\\

26 &
\rpmMetric{plan_energy_js} &
\(\displaystyle
\begin{aligned}
&\frac12\operatorname{KL}(\rho\|\omega)
+\frac12\operatorname{KL}(\eta\|\omega),\\
&\omega=\frac12(\rho+\eta)
\end{aligned}
\)
\\

27 &
\rpmMetric{sign_flip_fraction} &
\(\displaystyle
\frac{1}{Ld}\sum_{i,j}
\mathbf{1}[U_{ij}V_{ij}<0]
\)
\\

28 &
\rpmMetric{first_difference_nmse} &
\(\displaystyle
\frac{1}{(L-1)d}
\sum_{i=0}^{L-2}\sum_j
(\Delta\widetilde U_{ij}
-\Delta\widetilde V_{ij})^2
\)
\\

29 &
\rpmMetric{second_difference_nmse} &
\(\displaystyle
\frac{1}{(L-2)d}
\sum_{i=0}^{L-3}\sum_j
(\Delta^2\widetilde U_{ij}
-\Delta^2\widetilde V_{ij})^2
\)
\\

30 &
\rpmMetric{relative_total_variation_gap} &
\(\displaystyle
\frac{
|\operatorname{TV}(\widetilde U)
-\operatorname{TV}(\widetilde V)|
}{
\tfrac12[
\operatorname{TV}(\widetilde U)
+\operatorname{TV}(\widetilde V)
]+\epsilon
}
\)
\\

31 &
\rpmMetric{cumulative_mean_drift} &
\(\displaystyle
\frac{1}{L}\sum_{i=0}^{L-1}
\left\|
\sum_{r=0}^{i}\widetilde E_r
\right\|_2
\)
\\

32 &
\rpmMetric{cumulative_endpoint_drift} &
\(\displaystyle
\left\|\sum_i\widetilde E_i\right\|_2
\)
\\

33 &
\rpmMetric{overlap_energy_distance} &
\(\displaystyle
\begin{aligned}
\Biggl[
&\frac{2}{L^2}\sum_{i,r}
\|\widetilde U_i-\widetilde V_r\|_2\\
-&\frac{1}{L^2}\sum_{i,r}
\|\widetilde U_i-\widetilde U_r\|_2\\
-&\frac{1}{L^2}\sum_{i,r}
\|\widetilde V_i-\widetilde V_r\|_2
\Biggr]_+
\end{aligned}
\)
\\

34 &
\rpmMetric{mean_dimension_wasserstein1} &
\(\displaystyle
\frac{1}{Ld}\sum_j\sum_i
|\widetilde U_{(i)j}-\widetilde V_{(i)j}|
\)
\\

35 &
\rpmMetric{overlap_rbf_mmd2_multiscale} &
\(\displaystyle
\left[
\frac{1}{|\mathcal B|}
\sum_{b\in\mathcal B}
\operatorname{MMD}_b^2(\widetilde U,\widetilde V)
\right]_+
\)
\\

\end{longtable}
\endgroup

\paragraph{Auxiliary definitions.}
Here \(r_j\) and \(\sigma_j^{\mathrm{cal}}\) are the
stabilized training-action ranges and standard deviations
used for metric calibration.
The matrix \(\Sigma_a^\dagger\) is the pseudoinverse
of the regularized training-action covariance matrix.

The Huber penalty is
\[
h_1(x)=
\begin{cases}
\tfrac12x^2, & |x|\leq1,\\
|x|-\tfrac12, & |x|>1.
\end{cases}
\]
For a sequence \(Z\) of length \(L\), temporal differences
and total variation are
\[
\begin{aligned}
\Delta Z_i
&=Z_{i+1}-Z_i,
&& i=0,\ldots,L-2,\\
\Delta^2 Z_i
&=Z_{i+2}-2Z_{i+1}+Z_i,
&& i=0,\ldots,L-3,\\
\operatorname{TV}(Z)
&=\sum_{i=0}^{L-2}\|Z_{i+1}-Z_i\|_2.
\end{aligned}
\]
The notation \([x]_+=\max(0,x)\) denotes nonnegative clipping.

\paragraph{Plan-energy and distributional measures.}
The plan-energy distributions are
\[
\rho_n=
\frac{
|(\operatorname{vec}\widetilde U)_n|+\epsilon_t
}{
\sum_m\bigl(
|(\operatorname{vec}\widetilde U)_m|+\epsilon_t
\bigr)
},
\qquad
\eta_n=
\frac{
|(\operatorname{vec}\widetilde V)_n|+\epsilon_t
}{
\sum_m\bigl(
|(\operatorname{vec}\widetilde V)_m|+\epsilon_t
\bigr)
}.
\]
For Wasserstein-1, parenthesized indices denote independently
sorted values within each action component.
The multiscale MMD uses \(\mathcal B=\{0.5,1,2\}\):
\[
\operatorname{MMD}_b^2(X,Y)
=
\frac{1}{L^2}\sum_{i,r}
\left[
k_b(X_i,X_r)+k_b(Y_i,Y_r)-2k_b(X_i,Y_r)
\right],
\]
where
\[
k_b(x,y)=
\exp\!\left(-\frac{\|x-y\|_2^2}{2b^2}\right).
\]
The within-sequence sums include diagonal terms.

\paragraph{Action-component measures.}
The vectors \(u_i,v_i\) are the translation components of
\(U_i,V_i\). Their cosine distance is
\[
c(u,v)=
1-\frac{u^\top v}
{(\|u\|_2+\epsilon)(\|v\|_2+\epsilon)}.
\]
Pairs with both vectors below the translation dead-zone
threshold have distance zero.
Pairs with only one vector below the threshold are excluded,
and \(\mathcal I\) denotes the remaining positions.

The quantities \(t_i^U,t_i^V\) and
\(\mathcal R_i^U,\mathcal R_i^V\) are
controller-preprocessed translation increments and rotation
matrices. The quantities \(g_i^U,g_i^V\) are gripper commands.
Their calibration scales are \(\sigma_p,\sigma_R,s_g\).
Rotation disagreement uses
\[
d_{\mathrm{SO}(3)}(\mathcal R,\mathcal S)
=
\arccos\!\left(
\operatorname{clip}_{[-1,1]}
\frac{\operatorname{tr}(\mathcal R^\top\mathcal S)-1}{2}
\right).
\]

\paragraph{Temporal measures.}
For the single-task screening, EMA deviation uses
\(\alpha=0.3\).
CUSUM uses \(\kappa=0.5\), with reference mean
\(\mu_0^s\) and stabilized standard deviation
\(\sigma_0^s\) computed from the first five finite
values of \(s_q\).
The range-normalized EMA is initialized with its first
finite \(A_q^{\mathrm{ACC}}\).
These metric definitions use
\(\epsilon=10^{-12}\) and \(\epsilon_t=10^{-8}\).

\end{document}